\documentclass{article}

\usepackage[nonatbib,final]{neurips_2020}

\usepackage[utf8]{inputenc} 
\usepackage[T1]{fontenc}    
\usepackage{hyperref}       
\usepackage{url}            
\usepackage{booktabs}       
\usepackage{amsfonts}       
\usepackage{nicefrac}       
\usepackage{microtype}      
\usepackage{booktabs}
\usepackage{multirow}
\usepackage[normalem]{ulem}
\usepackage{adjustbox}
\usepackage{xspace}
\usepackage{pifont}
\usepackage{amsthm}
\usepackage{amsmath} 
\usepackage{dsfont}
\usepackage{graphicx}
\usepackage{caption}
\usepackage{subcaption}
\usepackage{xcolor}
\usepackage{wrapfig}
\usepackage{tikz}
\usepackage{algorithm}
\usepackage[noend]{algpseudocode}
\usetikzlibrary{arrows,positioning,shapes,calc}
\tikzset{
  treenode/.style = {shape=rectangle, rounded corners,
                     draw, align=center, fill = blue!15!white}, 
  treenode2/.style = {shape=rectangle, rounded corners,
                     draw, align=center, fill = red!15!white}, 
  root/.style     = {rectangle, font=\scriptsize\bf, draw, align=center, fill=red!15!white},
  env/.style      = {treenode, font=\scriptsize\bf},
  env2/.style     = {treenode, draw=black,font=\scriptsize\bf ,fill=blue!15!white, very thick},
  env3/.style     = {treenode, font=\scriptsize\bf ,fill=green!15!white, very thick},
  newenv/.style     = {draw=none, font=\scriptsize\bf, align=center},
  whiteenv/.style     = {draw=none, font=\scriptsize, align=center},
  level2/.style     ={newenv, level distance=10cm, sibling distance=0.5em},
}

\DeclareMathOperator*{\argmax}{arg\,max}
\DeclareMathOperator*{\argmin}{arg\,min}

\newcommand{\ac}{SupSup\xspace}

\newif\ifcomments

\commentsfalse

\ifcomments
\newcommand{\comments}[1]{#1}
\else
\newcommand{\comments}[1]{}
\fi

\newcommand{\rosanne}[1]{\comments{\textcolor{magenta}{[rosanne: #1]}}}

\definecolor{muchlater}{rgb}{0.7,1,.7}
\newcommand{\muchlater}[1]{\comments{\textcolor{muchlater}{[Much Later: #1]}}}

\newcommand{\casename}[1]{\ensuremath{\mathsf{#1}}\xspace}

\newcommand{\figlabel}[1]{\label{fig:#1}}
\newcommand{\figref}[1]{Figure~\ref{fig:#1}}

\newcommand{\secref}[1]{Section~\ref{sec:#1}}

\newcommand{\tabref}[1]{Table~\ref{tab:#1}}

\title{Supermasks in Superposition}

\author{%
  Mitchell Wortsman\thanks{Equal contribution. $^\dagger$Also affiliated with the University of Washington. Code available at \url{https://github.com/RAIVNLab/supsup} and correspondence to \texttt{\{mitchnw,ramanv\}@cs.washington.edu}.} \\
    University of Washington \\
    \And
      Vivek Ramanujan$^*$ \\
    Allen Institute for AI \\
        \And
      Rosanne Liu \\
    ML Collective \\
        \And
      Aniruddha Kembhavi$^\dagger$ \\
    Allen Institute for AI \\
        \And
      Mohammad Rastegari \\
    University of Washington \\
            \And
      Jason Yosinski \\
    ML Collective \\
        \And
      Ali Farhadi \\
    University of Washington \\
}
\begin{document}

\theoremstyle{plain}
\newtheorem{theorem}{Theorem}[section]
\newtheorem{corollary}{Corollary}[theorem]
\newtheorem{lemma}[theorem]{Lemma}
\newtheorem{proposition}[theorem]{Proposition}

\theoremstyle{definition}
\newtheorem{definition}[theorem]{Definition}
\newtheorem{remark}[theorem]{Remark}

\newcommand{\round}[1]{\left( #1 \right)}
\newcommand{\curly}[1]{\left\lbrace #1 \right\rbrace}
\newcommand{\squarebrack}[1]{\left\lbrack #1 \right\rbrack}

\newcommand{\sumi}[2]{\sum\limits_{i=#1}^{#2}}
\newcommand{\sumj}[2]{\sum\limits_{j=#1}^{#2}}
\newcommand{\sumk}[2]{\sum\limits_{k=#1}^{#2}}
\newcommand{\sump}[2]{\sum\limits_{p=#1}^{#2}}
\newcommand{\suml}[2]{\sum\limits_{l=#1}^{#2}}
\newcommand{\sumn}[2]{\sum\limits_{n=#1}^{#2}}
\newcommand{\summ}[2]{\sum\limits_{m=#1}^{#2}}
\newcommand{\sumt}[2]{\sum\limits_{t=#1}^{#2}}

\newcommand{\Sum}{\sum_{i = 1}^{n}}
\newcommand{\Sumi}[1]{\sum\limits_{i = 1}^{#1}}
\newcommand{\Sumt}[1]{\sum\limits_{t = 1}^{#1}}

\newcommand{\abs}[1]{\left\lvert #1 \right\rvert}
\newcommand{\norm}[2]{\left\lVert#2\right\rVert_{#1}}
\newcommand{\esqnorm}[1]{\left\lVert#1\right\rVert_2^2}
\newcommand{\enorm}[1]{\left\lVert#1\right\rVert_2}
\newcommand{\infnorm}[1]{\left\lVert#1\right\rVert_\infty}
\newcommand{\opnorm}[1]{\left\lVert#1\right\rVert_\text{op}}
\newcommand{\normF}[1]{\left\lVert#1\right\rVert_{\text{F}}}
\newcommand{\inner}[1]{\left\langle#1\right\rangle}
\newcommand{\ceil}[1]{\left\lceil#1\right\rceil}
\newcommand{\floor}[1]{\left\lfloor#1\right\rfloor}

\newcommand{\zero}{\mathbf{0}}
\newcommand{\one}{\mathbf{1}}

\newcommand{\avec}{\mathbf{a}}
\newcommand{\bvec}{\mathbf{b}}
\newcommand{\cvec}{\mathbf{c}}
\newcommand{\dvec}{\mathbf{d}}
\newcommand{\e}{\mathbf{e}}
\newcommand{\f}{\mathbf{f}}
\newcommand{\g}{\mathbf{g}}
\newcommand{\h}{\mathbf{h}}
\newcommand{\ivec}{\mathbf{i}}
\newcommand{\jvec}{\mathbf{j}}
\newcommand{\kvec}{\mathbf{k}}
\newcommand{\lvec}{\mathbf{l}}
\newcommand{\m}{\mathbf{m}}
\newcommand{\n}{\mathbf{n}}
\newcommand{\ovec}{\mathbf{o}}
\newcommand{\p}{\mathbf{p}}
\newcommand{\q}{\mathbf{q}}
\newcommand{\rvec}{\mathbf{r}}
\newcommand{\s}{\mathbf{s}}
\newcommand{\tvec}{\mathbf{t}}
\newcommand{\uvec}{\mathbf{u}}
\newcommand{\vvec}{\mathbf{v}}
\newcommand{\w}{\mathbf{w}}
\newcommand{\x}{\mathbf{x}}
\newcommand{\y}{\mathbf{y}}
\newcommand{\z}{\mathbf{z}}

\newcommand{\A}{\mathbf{A}}
\newcommand{\B}{\mathbf{B}}
\newcommand{\C}{\mathbf{C}}
\newcommand{\D}{\mathbf{D}}
\newcommand{\Emat}{\mathbf{E}}
\newcommand{\F}{\mathbf{F}}
\newcommand{\G}{\mathbf{G}}
\newcommand{\Hmat}{\mathbf{H}}
\newcommand{\I}{\mathbf{I}}
\newcommand{\J}{\mathbf{J}}
\newcommand{\K}{\mathbf{K}}
\newcommand{\Lmat}{\mathbf{L}}
\newcommand{\M}{\mathbf{M}}
\newcommand{\N}{\mathbf{N}}
\newcommand{\Omat}{\mathbf{O}}
\newcommand{\Pmat}{\mathbf{P}}
\newcommand{\Q}{\mathbf{Q}}
\newcommand{\Rmat}{\mathbf{R}}
\newcommand{\Smat}{\mathbf{S}}
\newcommand{\T}{\mathbf{T}}
\newcommand{\U}{\mathbf{U}}
\newcommand{\V}{\mathbf{V}}
\newcommand{\W}{\mathbf{W}}
\newcommand{\X}{\mathbf{X}}
\newcommand{\Y}{\mathbf{Y}}
\newcommand{\Z}{\mathbf{Z}}

\newcommand{\SIGMA}{\mathbf{\Sigma}}
\newcommand{\LAMBDA}{\mathbf{\Lambda}}

\newcommand{\Acal}{\mathcal{A}}
\newcommand{\Bcal}{\mathcal{B}}
\newcommand{\Ccal}{\mathcal{C}}
\newcommand{\Dcal}{\mathcal{D}}
\newcommand{\Ecal}{\mathcal{E}}
\newcommand{\Fcal}{\mathcal{F}}
\newcommand{\Gcal}{\mathcal{G}}
\newcommand{\Hcal}{\mathcal{H}}
\newcommand{\Ical}{\mathcal{I}}
\newcommand{\Jcal}{\mathcal{J}}
\newcommand{\Kcal}{\mathcal{K}}
\newcommand{\Lcal}{\mathcal{L}}
\newcommand{\Mcal}{\mathcal{M}}
\newcommand{\Ncal}{\mathcal{N}}
\newcommand{\Ocal}{\mathcal{O}}
\newcommand{\Pcal}{\mathcal{P}}
\newcommand{\Qcal}{\mathcal{Q}}
\newcommand{\Rcal}{\mathcal{R}}
\newcommand{\Scal}{\mathcal{S}}
\newcommand{\Tcal}{\mathcal{T}}
\newcommand{\Ucal}{\mathcal{U}}
\newcommand{\Vcal}{\mathcal{V}}
\newcommand{\Wcal}{\mathcal{W}}
\newcommand{\Xcal}{\mathcal{X}}
\newcommand{\Ycal}{\mathcal{Y}}
\newcommand{\Zcal}{\mathcal{Z}}

\newcommand{\alphavec}{\boldsymbol{\alpha}}
\newcommand{\betavec}{\boldsymbol{\beta}}
\newcommand{\gammavec}{\boldsymbol{\gamma}}
\newcommand{\deltavec}{\boldsymbol{\delta}}
\newcommand{\epsvec}{\boldsymbol{\epsilon}}
\newcommand{\etavec}{\boldsymbol{\eta}}
\newcommand{\nuvec}{\boldsymbol{\nu}}
\newcommand{\tauvec}{\boldsymbol{\tau}}
\newcommand{\rhovec}{\boldsymbol{\rho}}
\newcommand{\lmbda}{\boldsymbol{\lambda}}
\newcommand{\muvec}{\boldsymbol{\mu}}
\newcommand{\thetavec}{\boldsymbol{\theta}}

\newcommand{\BigO}[1]{\mathcal{O}\round{#1}}
\newcommand{\BigOmega}[1]{\Omega\round{#1}}

\newcommand{\R}{\mathbb{R}}
\newcommand{\Rd}[1]{\mathbb{R}^{#1}}
\newcommand{\Natural}{\mathbb{N}}
\newcommand{\Complex}{\mathbb{C}}
\newcommand{\Integer}{\mathbb{Z}}
\newcommand{\Rational}{\mathbb{Q}}

\newcommand{\E}[1]{\mathbb{E}\squarebrack{#1}}
\newcommand{\Exp}[2]{\mathbb{E}_{#1}\squarebrack{#2}}
\newcommand{\Prob}[1]{\mathds{P}\squarebrack{#1}}
\newcommand{\Probability}[2]{P_{#1}\curly{#2}}
\newcommand{\Var}[1]{\mathrm{Var}\squarebrack{#1}}
\newcommand{\Cov}[1]{\mathrm{Cov}\squarebrack{#1}}
\newcommand{\PR}[1]{\mathds{P}\round{#1}}

\newcommand{\inv}[1]{\frac{1}{#1}}
\newcommand{\indicator}[2]{\mathbbm{1}_{#1}\squarebrack{#2}}
\newcommand{\Tr}[1]{\text{Tr}\squarebrack{#1}}

\newcommand{\BOX}[1]{\fbox{\parbox{\linewidth}{\centering#1}}}
\newcommand{\textequal}[1]{\stackrel{#1}{=}}
\newcommand{\textleq}[1]{\stackrel{#1}{\leq}}
\newcommand{\textgeq}[1]{\stackrel{#1}{\geq}}
\newcommand{\defeq}{\vcentcolon=}

\newcommand{\dd}[2]{\frac{d #1}{d #2}}
\newcommand{\ddn}[3]{\frac{d^{#1} #2}{d #3^{#1}}}
\newcommand{\dodo}[2]{\frac{\partial #1}{\partial #2}}
\newcommand{\dodon}[3]{\frac{\partial^{#1} #2}{\partial {#3}^{#1}}}

\maketitle

\begin{abstract}
We present the Supermasks in Superposition (SupSup) model, capable of sequentially learning thousands of tasks without catastrophic forgetting. Our approach uses a randomly initialized, fixed base network and for each task finds a subnetwork (supermask) that achieves good performance. If task identity is given at test time, the correct subnetwork can be retrieved with minimal memory usage. If not provided, SupSup can infer the task using gradient-based optimization to find a linear superposition of learned supermasks which minimizes the output entropy. In practice we find that a single gradient step is often sufficient to identify the correct mask, even among 2500 tasks. We also showcase two promising extensions. First, SupSup models can be trained entirely without task identity information, as they may detect when they are uncertain about new data and allocate an additional supermask for the new training distribution. Finally the entire, growing set of supermasks can be stored in a constant-sized reservoir by implicitly storing them as attractors in a fixed-sized Hopfield network.
\end{abstract}

\section{Introduction}

Learning many different tasks sequentially without forgetting remains a notable challenge for neural networks \cite{thrun1998lifelong, zhao1996incremental, kirkpatrick2017overcoming}. If the weights of a neural network are trained on a new task, performance on previous tasks often degrades substantially \cite{mccloskey1989catastrophic,french1999catastrophic, goodfellow2013empirical}, a problem known as \emph{catastrophic forgetting}.
In this paper, we begin with the observation that catastrophic forgetting cannot occur if the weights of the network remain fixed and random. We leverage this to develop a flexible model capable of learning thousands of tasks: \textit{Supermasks in Superposition} (\ac). \ac, diagrammed in \figref{teaser}, is driven by two core ideas: \textbf{a)} the expressive power of untrained, randomly weighted subnetworks \cite{zhou2019deconstructing, ramanujan2019s}, and \textbf{b)} inference of task-identity as a gradient-based optimization problem.

\paragraph{a) The expressive power of subnetworks}
Neural networks may be overlaid with a binary mask that selectively keeps or removes each connection, producing a subnetwork.
The number of possible subnetworks is combinatorial in the number of parameters. Researchers have observed that the number of combinations is large enough 
that even within randomly weighted neural networks, there exist \emph{supermasks} that create corresponding subnetworks which achieve good performance on complex tasks. Zhou \textit{et al.} \cite{zhou2019deconstructing} and Ramanujan \textit{et al.} \cite{ramanujan2019s} present two algorithms for finding these supermasks while keeping the weights of the underlying network fixed and random.
SupSup scales to many tasks by finding for each task a supermask atop a shared, untrained network.

\paragraph{b) Inference of task-identity as an optimization problem}
When task identity is unknown, $\text{\ac}$ can infer task identity to select the correct supermask.
Given data from task $j$, we aim to recover and use the supermask originally trained for task $j$. This supermask should exhibit a confident (\textit{i.e.} low entropy) output distribution when given data from task $j$ \cite{hendrycks2016baseline}, so we frame inference of task-identity as an optimization problem---find the convex combination of learned supermasks which minimizes the entropy of the output distribution.

In the rest of the paper we develop and evaluate \ac via the following contributions:
\begin{enumerate}
    \item We propose a new taxonomy of continual learning scenarios. We use it to embed and contextualize related work (\secref{cl}).
    \item When task identity (ID) is provided during train and test (later dubbed \casename{GG}), \ac is a natural extension of Mallya \textit{et al.} \cite{mallya2018piggyback}. By using a randomly weighted backbone and controlling mask sparsity, \ac surpasses recent baselines on SplitImageNet \cite{wen2020batchensemble} while requiring less storage and time costs (\secref{S1}).
    \item When task ID is provided during train but not test (later dubbed \casename{GN}), \ac outperforms recent methods that require task ID \cite{mnist, kirkpatrick2017overcoming, cheung2019superposition}, scaling to 2500 permutations of MNIST without forgetting. For these uniform tasks, ID can be inferred with a single gradient computation (\secref{S23}). 
    \item When task identities are not provided at all (later dubbed \casename{NNs}), \ac can even infer task boundaries and allocate new supermasks as needed (\secref{S4}).
    \item We introduce an extension to the basic \ac algorithm that stores supermasks implicitly as attractors in a fixed-size Hopfield network \cite{hopfield1982neural} (\secref{hop}).    
    \item Finally, we empirically show that the simple trick of adding \emph{superfluous neurons} results in more accurate task inference (\secref{s-neuron}).
\end{enumerate}

\begin{figure}[t]
    \centering
    \includegraphics[width=\textwidth]{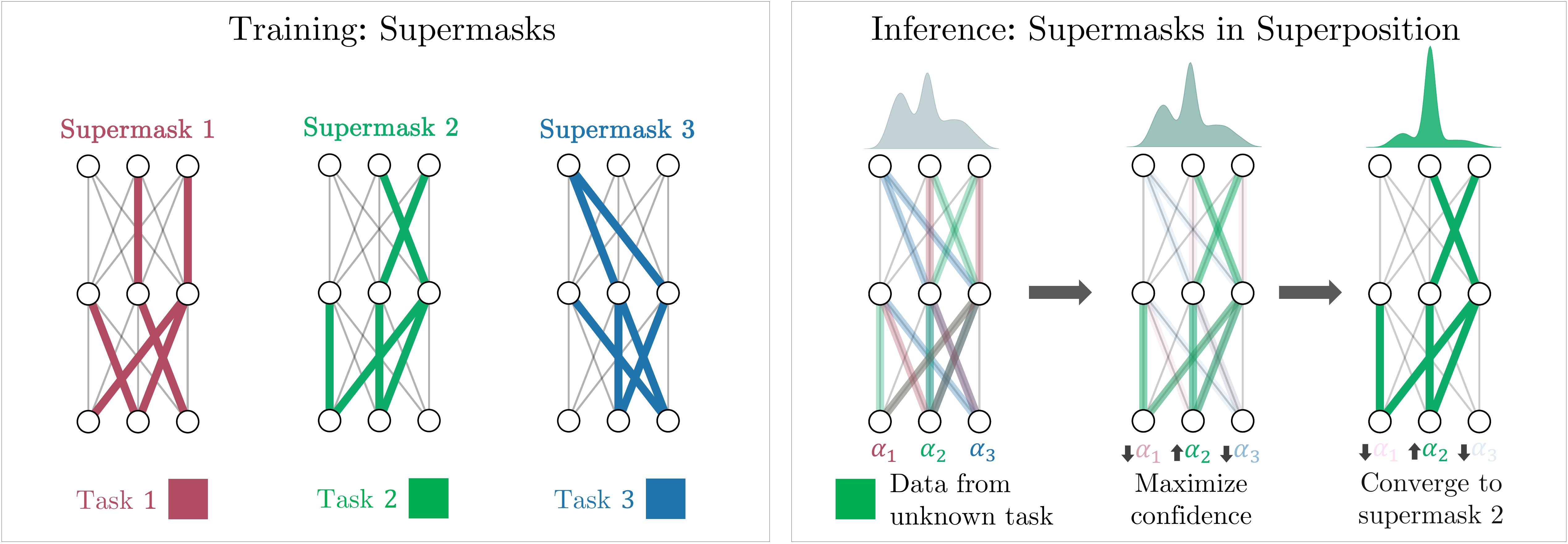}
    \caption{\textbf{(left)} During training \ac learns a separate supermask (subnetwork) for each task. \textbf{(right)} At inference time, \ac can infer task identity by superimposing all supermasks, each weighted by an $\alpha_i$, and using gradients to maximize confidence.}
    \vspace*{-2ex}
    \figlabel{teaser}
\end{figure}

\vspace*{-1ex}
\section{Continual Learning Scenarios and Related Work}
\label{sec:cl}
\vspace*{-1ex}

In continual learning, a model aims to solve a number of tasks sequentially \cite{thrun1998lifelong, zhao1996incremental} without catastrophic forgetting \cite{french1999catastrophic, kirkpatrick2017overcoming, mccloskey1989catastrophic}.
Although numerous approaches have been proposed in the context of continual learning, there lacks a convention of  scenarios in which methods are trained and evaluated~\cite{van2019three}. The key identifiers of scenarios include: \textbf{1)} whether task identity is provided during training, \textbf{2)} provided during inference, \textbf{3)} whether class labels are shared during evaluation, 
and \textbf{4)} whether the overall task space is discrete or continuous.
This results in an exhaustive set of 16 possibilities, many of which are invalid or uninteresting. For example, 
if task identity is never provided in training, providing it in inference is no longer helpful.
To that end, we highlight four applicable scenarios, each with a further breakdown of discrete vs. continuous, when applicable, as shown in~\tabref{scenarios}.

We decompose continual learning scenarios via a three-letter taxonomy that explicitly addresses the three most critical scenario variations. 
The first two letters specify whether task identity is given during training (\casename{G} if given, \casename{N} if not) and during inference (\casename{G} if given, \casename{N} if not).
The third letter specifies a subtle but important distinction: whether labels are shared (\casename{s}) across tasks or not (\casename{u}).
In the unshared case, the model must predict both the correct task ID and the correct class within that task. In the shared case, the model need only predict the correct, shared label across tasks, so it need not represent or predict which task the data came from.
For example, when learning 5 permutations of MNIST in the \casename{GN} scenario (task IDs given during train but not test), a shared label \casename{GNs} scenario will evaluate the model on the correct predicted label across 10 possibilities, while in the unshared \casename{GNu} case the model must predict across 50 possibilities, a more difficult problem.

A full expansion of possibilities entails both \casename{GGs} and \casename{GGu}, but as \casename{s} and \casename{u} describe only model \emph{evaluation}, any model capable of predicting shared labels can predict unshared equally well using the provided task ID at test time. Thus these cases are equivalent, and we designate both \casename{GG}.
Moreover,
the \casename{NNu} scenario is invalid because unseen labels signal the presence of a new task (the ``labels trick'' in \cite{zeno2018task}), making the scenario actually \casename{GNu}, and so we consider only the shared label case \casename{NNs}.

We leave out the discrete vs. continuous distinction as most research efforts operate within one framework or the other, and the taxonomy applies equivalently to discrete domains with integer ``Task IDs'' as to continue domains with ``Task Embedding'' or ``Task Context'' vectors. The remainder of this paper follows the majority of extant literature in focusing on the case with discrete task boundaries (see e.g. \cite{ zeno2018task} for progress in the continuous scenario).
Equipped with this taxonomy, we review three existing approaches for continual learning.

\begin{table}[t]
  \caption{Overview of different Continual Learning scenarios. We suggest scenario names that provide an intuitive understanding of the variations in training, inference, and evaluation, while allowing a full coverage of the scenarios previously defined in~\cite{van2019three} and~\cite{zeno2018task}. See text for more complete description.
    \muchlater{can we cover regression, unsupervised learning, and RL too?}
}
\vspace{-1em}
\label{tab:scenarios}
\centering
\begin{adjustbox}{width=1\textwidth}
\begin{tabular}{@{}l l l l@{}}
\toprule
\multirow{2}{*}{Scenario} &
  \multirow{2}{*}{Description} &
  \multirow{2}{*}{\begin{tabular}[c]{@{}l@{}}Task space discreet\\ or continuous?\end{tabular}} &
  \multirow{2}{*}{\begin{tabular}[c]{@{}l@{}}Example methods / \\ task names used\end{tabular}} \\
 &
   &
   &
   \\ \midrule
  \casename{GG} &
  Task \textbf{G}iven during train and \textbf{G}iven during inference &
  Either &

  \begin{tabular}[c]{@{}l@{}}PNN~\cite{rusu2016progressive}, BatchE~\cite{wen2020batchensemble}, PSP~\cite{cheung2019superposition}, ``Task learning''~\cite{zeno2018task}, ``Task-IL''~\cite{van2019three} \end{tabular} \\ \midrule
    \casename{GNs} & Task \textbf{G}iven during train, \textbf{N}ot inference; \textbf{s}hared labels      & Either          & EWC~\cite{kirkpatrick2017overcoming}, SI \cite{zenke2017continual}, ``Domain learning''~\cite{zeno2018task}, ``Domain-IL''~\cite{van2019three} \\ \midrule
    \casename{GNu} & Task \textbf{G}iven during train, \textbf{N}ot inference; \textbf{u}nshared labels & Discrete only &    ``Class learning''~\cite{zeno2018task}, ``Class-IL''~\cite{van2019three}                                                              \\ \midrule
\casename{NNs} &
  Task \textbf{N}ot given during train \textbf{N}or inference; \textbf{s}hared labels &
  Either &
  BGD, ``Continuous/discrete task agnostic learning''~\cite{zeno2018task} \\ \bottomrule
\end{tabular}
\end{adjustbox}
\vspace{-1.69em}
\end{table}

\textbf{(1) Regularization based methods }
Methods like Elastic Weight Consolidation (EWC) \cite{kirkpatrick2017overcoming} and Synaptic Intelligence (SI) \cite{zenke2017continual} penalize the movement of parameters that are important for solving previous tasks 
in order to mitigate catastrophic forgetting.
Measures of parameter importance vary; e.g. EWC uses the Fisher Information matrix \cite{pascanu2013revisiting}. These methods operate in the \casename{GNs} scenario (\tabref{scenarios}).
Regularization approaches ameliorate but do not exactly eliminate catastrophic forgetting.

\textbf{(2) Using exemplars, replay, or generative models }
These methods aim to explicitly or implicitly (with generative models) capture data from previous tasks. For instance,~\cite{rebuffi2017icarl} performs classification based on the nearest-mean-of-examplars in a feature space. Additionally, \cite{lopez2017gradient, chaudhry2018efficient} prevent the model from increasing loss on examples from previous tasks while \cite{rolnick2019experience} and \cite{shin2017continual} respectively use memory buffers and generative models to replay past data.
Exact replay of the entire dataset can trivially eliminate catastrophic forgetting but at great time and memory cost. Generative approaches can reduce catastrophic forgetting, but generators are also susceptible to forgetting. Recently, \cite{Oswald2020Continual} successfully mitigate this obstacle by parameterizing a generator with a hypernetwork \cite{ha2016hypernetworks}.  

\textbf{(3) Task-specific model components }
Instead of modifying the learning objective or replaying data, various methods~\cite{rusu2016progressive,yoon2017lifelong, mallya2018packnet, mallya2018piggyback, masse2018alleviating, xu2018reinforced,cheung2019superposition, golkar2019continual, wen2020batchensemble} use different model components for different tasks. In Progressive Neural Networks (PNN), Dynamically Expandable Networks (DEN), and Reinforced Continual Learning (RCL) \cite{rusu2016progressive, yoon2017lifelong, xu2018reinforced}, the model is expanded for each new task. More efficiently, \cite{masse2018alleviating} fixes the network size and randomly assigns which nodes are active for a given task. In \cite{mallya2018packnet, golkar2019continual}, the weights of disjoint subnetworks are trained for each new task. Instead of learning the weights of the subnetwork, for each new task Mallya \textit{et al.} \cite{mallya2018piggyback} learn a binary mask that is applied to a network pretrained on ImageNet. 
Recently, Cheung \textit{et al.}~\cite{cheung2019superposition} superimpose many models into one by using different (and nearly orthogonal) contexts for each task. The task parameters can then be effectively retrieved using the correct task context.
Finally, BatchE~\cite{wen2020batchensemble} learns a shared weight matrix on the first task and learn only a rank-one elementwise scaling matrix for each subsequent task.

Our method falls into this final approach (3) as it introduces task-specific supermasks.
However, while all other methods in this category are limited to the \casename{GG} scenario, SupSup can be used to achieve compelling performance in \emph{all four scenarios}.
We compare primarily with BatchE~\cite{wen2020batchensemble} and Parameter Superposition (abbreviated PSP) \cite{cheung2019superposition} as they are recent and performative. BatchE requires very few additional parameters for each new task while achieving comparable performance to PNN and scaling to SplitImagenet. Moreover, PSP outperforms regularization based approaches like SI \cite{zenke2017continual}.
However, both BatchE \cite{wen2020batchensemble} and PSP \cite{cheung2019superposition} require task identity to use task-specific weights, so they can only operate in the \casename{GG} setting.

\vspace*{-1ex}
\section{Methods} \label{sec:method}
\vspace*{-1ex}

In this section, we detail how \ac leverages supermasks to learn thousands of sequential tasks without forgetting. We begin with easier settings where task identity is given and gradually move to more challenging scenarios where task identity is unavailable.

\subsection{Preliminaries}

In a standard $\ell$-way classification task, inputs $\x$ are mapped to a distribution $\p$ over output neurons $\{1,...,\ell\}$. 
We consider the general case where $\p = f(\x, W)$ for a neural network $f$ parameterized by $W$ and trained with a cross-entropy loss. In continual learning classification settings we have $k$ different $\ell$-way classification tasks and the input size remains constant across tasks\footnote{In practice the tasks do not all need to be $\ell$-way --- output layers can be padded until all have the same size.}.

Zhou \textit{et al.} \cite{zhou2019deconstructing} demonstrate that a trained binary mask (supermask) $M$ can be applied to a randomly weighted neural network, resulting in a subnetwork with good performance. As further explored by Ramanujan \textit{et al.} \cite{ramanujan2019s}, supermasks can be trained at similar compute cost to training weights while achieving performance competitive with weight training.

With supermasks, outputs are given by $\p = f\round{\x, W \odot M}$ where $\odot$ denotes an elementwise product. $W$ is kept frozen at its initialization: bias terms are $\zero$ and other parameters in $W$  are $\pm c$ with equal probability and $c$ is the standard deviation of the corresponding Kaiming normal distribution \cite{he2015delving}. This initialization is referred to as \textit{signed Kaiming constant} by \cite{ramanujan2019s} and the constant $c$ may be different for each layer. For completeness we detail the Edge-Popup algorithm for training supermasks \cite{ramanujan2019s} in Section~\ref{sec:supermask-training} of the appendix.

\subsection{Scenario \casename{GG}: Task Identity Information Given During Train and Inference} \label{sec:S1}
\begin{figure}[t]
\begin{subfigure}{.46\textwidth}
\small
\vspace{-1.9em}
\begin{tabular}{lcr}
\toprule
          Algorithm &  Avg Top 1 &   Bytes \\
                 &  Accuracy (\%) &    \\
\midrule
                   Upper Bound &               92.55 &  10222.81M\\\midrule
  &               89.58 &    195.18M \\
\ac (\casename{GG}) &               88.68 &    100.98M \\
  &               86.37 &     65.50M \\\midrule
BatchE ($\ensuremath{\mathsf{GG}}$) &               81.50 &    124.99M \\
Single Model & - & 102.23M\\
\bottomrule
\end{tabular}
    \label{tab:splitimagenet}
\end{subfigure}
\begin{subfigure}{.54\textwidth}
    \centering
    \includegraphics[width=\textwidth]{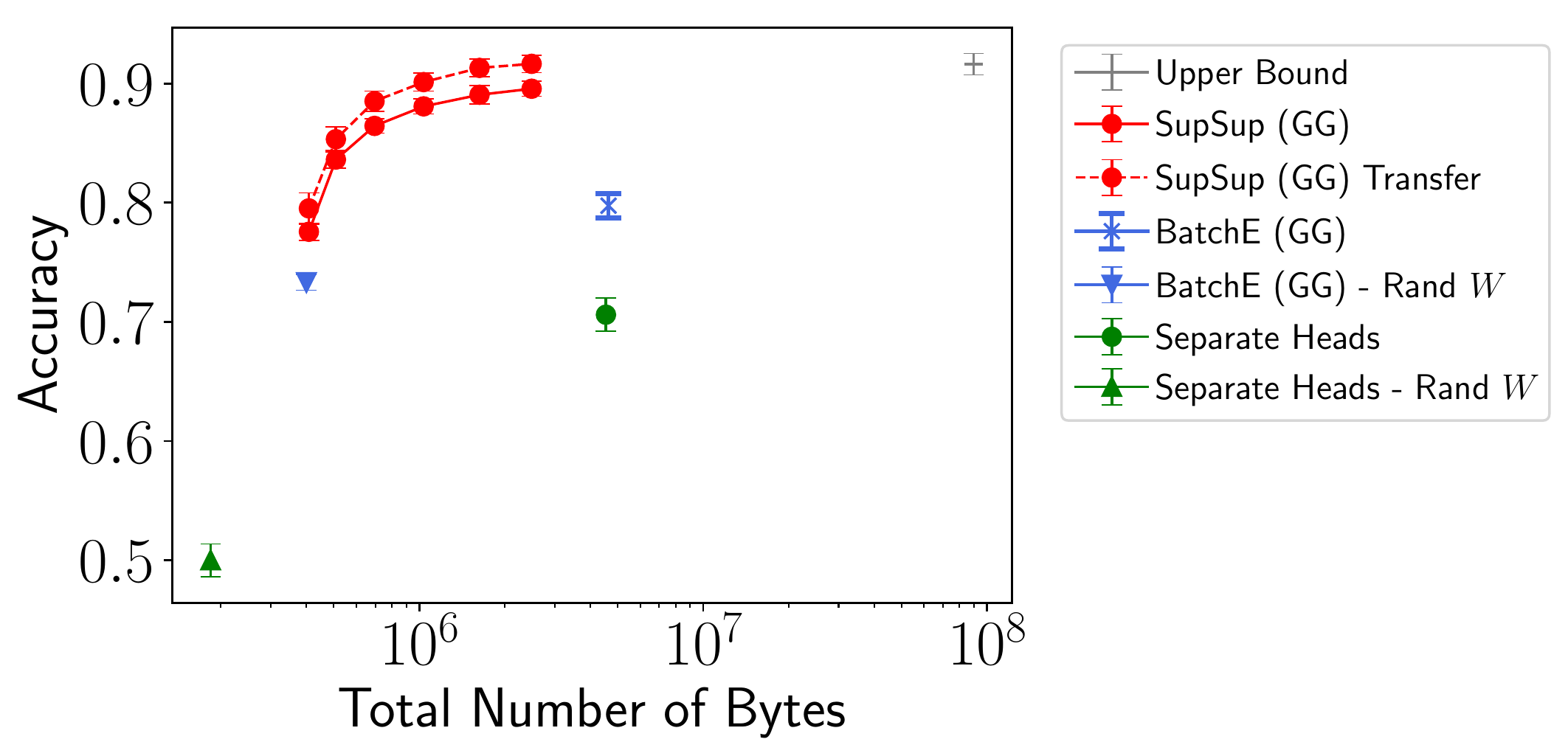}
    \label{fig:splitcifar}
    \end{subfigure}
    \vspace{-1.2em}
    \caption{\textbf{(left)} \textbf{SplitImagenet} performance in Scenario \casename{GG}. \ac approaches upper bound performance with significantly fewer bytes. \textbf{(right)} \textbf{SplitCIFAR100} performance in Scenario \casename{GG} shown as mean and standard deviation over 5 seed and splits. \ac outperforms similar size baselines and benefits from \textit{transfer}.}
    \vspace{-1.2em}
    \label{fig:t1}
\end{figure}

When task identity is known during training we can learn a binary mask $M^i$ per task. $M^i$ are the only parameters learned as the weights remain fixed. Given data from task $i$, outputs are computed as
\begin{equation} \label{eq:known-task-id}
    \p = f\round{\x, W \odot M^i}
\end{equation}
For each new task we can either initialize a new supermask randomly, or use a running mean of all supermasks learned so far.
During inference for task $i$ we then use $M^i$. \autoref{fig:t1} illustrates that in this scenario \ac outperforms a number of baselines in accuracy 
on both SplitCIFAR100 and SplitImageNet 
while requiring fewer bytes to store. Experiment details are in \secref{exp-s1}.

\subsection{Scenarios \casename{GNs} \& \casename{GNu} : Task Identity Information Given During Train Only} \label{sec:S23}
We now consider the case where input data comes from task $j$, but this task information is unknown to the model at inference time. During training we proceed exactly as in Scenario \casename{GG}, obtaining $k$ learned supermasks. During inference, we aim to infer task identity---correctly detect that the data belongs to task $j$---and select the corresponding supermask $M^j$. 

The \ac procedure for task ID inference is as follows: first we associate each of the $k$ learned supermasks $M^i$ with an coefficient $\alpha_i \in [0,1]$, initially set to $1/k$. Each $\alpha_i$ can be interpreted as the ``belief'' that supermask $M^i$ is the correct mask (equivalently the belief that the current unknown task is task $i$). The model's output is then be computed with a weighted superposition of all learned masks: 
\begin{equation} \label{eq:sup}
    \p(\alpha) = f\round{\x, W \odot \round{ \sum_{i=1}^k \alpha_i M^i}}.
\end{equation}

\hspace{-.5em}The correct mask $M^j$ should produce a confident, low-entropy output \cite{hendrycks2016baseline}. Therefore, to recover the correct mask we find the coefficients $\alpha$ which minimize the output entropy $\Hcal$ of $\p(\alpha)$. One option is to perform gradient descent on $\alpha$ via 
\begin{equation}\alpha \gets \alpha - \eta \nabla_\alpha \Hcal\round{\p \round{\alpha}}
\end{equation}
where $\eta$ is the step size, and $\alpha$s are re-normalized to sum to one after each update. Another option is to try each mask individually and pick the one with the lowest entropy output requiring $k$ forward passes. However, we want an optimization method with fixed sub-linear run time (w.r.t. the number of tasks $k$) which leads $\alpha$ to a corner of the probability simplex --- \textit{i.e.} $\alpha$ is 0 everywhere except for a single 1. We can then take the nonzero index to be the inferred task. To this end we consider the \textbf{One-Shot} and \textbf{Binary} algorithms.

\textbf{One-Shot:}
The task is inferred using a single gradient. Specifically, the inferred task is given by
\begin{equation} \label{eq:oneshot}
    \argmax_i  \round{- \frac{ \partial \Hcal\round{\p \round{\alpha}}}{\partial \alpha_i}}
\end{equation}
as entropy is decreasing maximally in this coordinate. This algorithms corresponds to one step of the Frank-Wolfe algorithm~\cite{frank1956algorithm}, or one-step of gradient descent followed by softmax re-normalization with the step size $\eta$ approaching $\infty$. Unless noted otherwise, $\x$ is a single image and not a batch.

\textbf{Binary:}
Resembling binary search, we infer task identity using an algorithm with $\log k$ steps. At each step we rule out half the tasks---the tasks corresponding to entries in the bottom half of $ - \nabla_\alpha \Hcal\round{\p \round{\alpha}}$. These are the coordinates in which entropy is minimally decreasing. A task $i$ is ruled out by setting $\alpha_i$ to zero and at each step we re-normalize the remaining entries in $\alpha$ so that they sum to one. Pseudo-code for both algorithms may be found in Section~\ref{sec:pseudo} of the appendix.

Once the task is inferred the corresponding mask can be used as in Equation~\ref{eq:known-task-id} to obtain class probabilities $\p$. In both Scenario \casename{GNs} and \casename{GNu} the class probabilities $\p$ are returned. In \casename{GNu}, $\p$ forms a distribution over the classes corresponding to the inferred task. Experiments solving thousands of tasks are detailed in \secref{exp-s23}.

\subsection{Scenario \casename{NNs}: No Task Identity During Training or Inference} \label{sec:S4}
Task inference algorithms from Scenario \casename{GN} enable the extension of \ac to Scenario \casename{NNs}, where task identity is entirely unknown (even during training). If \ac is uncertain about the current task identity, it is likely that the data do not belong to any task seen so far. When this occurs a new supermask is allocated, and $k$ (the number of tasks learned so far) is incremented. 

We consider the \textbf{One-Shot} algorithm and say that \ac is uncertain when performing task identity inference if $\nu = \texttt{softmax}\round{-\nabla_\alpha \Hcal \round{\p\round{\alpha}}}$ is approximately uniform. Specifically, if $k \max_i \nu_i < 1 + \epsilon $ a new mask is allocated and $k$ is incremented. Otherwise mask $\argmax_i \nu_i$ is used, which corresponds to Equation~\ref{eq:oneshot}. We conduct experiments on learning up to 2500 tasks entirely without any task information, detailed in \secref{exp-s4}. \autoref{fig:long-zoom} shows that \ac in Scenario \casename{NNs} achieves comparable performance even to Scenario \casename{GNu}.

\subsection{Beyond Linear Memory Dependence} \label{sec:hop} \label{sec:beyond-linear-mem}
Hopfield networks \cite{hopfield1982neural} implicitly encode a series of binary strings $\z^i \in \{-1,1\}^d$ with an associated energy function $E_\Psi(\z) = \sum_{uv} \Psi_{uv}\z_u\z_v$. Each $\z^i$ is a minima of $E_\Psi$, and can be recovered with gradient descent. $\Psi \in \R^{d \times d}$ is initially $\zero$, and to encode a new string $z^i$, $\Psi \gets \Psi + \frac{1}{d}\z^i {\z^i}^\top $.

We now consider implicitly encoding the masks in a fixed-size Hopfield network $\Psi$ for Scenario \casename{GNu}. For a new task $i$ a new mask is learned. After training on task $i$, this mask will be stored as an attractor in a fixed size Hopfield network. Given new data during inference we perform gradient descent on the Hopfield energy $E_\Psi$ with the output entropy $\Hcal$ to learn a new mask $\m$. Minimizing $E_\Psi$ will hopefully push $\m$ towards a mask learned during training while $\Hcal$ will push $\m$ to be the correct mask. 
As $\Psi$ is quadratic in mask size, we will not mask the parameters $W$. Instead we mask the output of every layer except the last, $\textit{e.g.}$ a network with one hidden layer and mask $\m$ is given by
\begin{align}
    f(\x, \m, W) = \texttt{softmax}\round{W_2^\top \round{\m \odot \sigma\round{W_1^\top \x }} }
\end{align}
for nonlinearity $\sigma$. The Hopfield network will then be a similar size as the base neural network. We refer to this method as Hop\ac and provide additional details in Section~\ref{sec:hop-extended}.\rosanne{Can we quickly summarize the result here?}
\rosanne{NOTE: since for NeurIPS SI and main paper are separate (reader cannot click this can be directed there immediately), it is better to add "In Appendix" or "In Supplementary Information" wherever possible.}

\subsection{Superfluous Neurons \& an Entropy Alternative} \label{sec:s-neuron}
Similar to previous methods \cite{van2019three}, Hop\ac requires $\ell k$ output neurons in Scenario \casename{GNu}. \ac, however, is performing $\ell k$-way classification without $\ell k$ output neurons. Given data during inference \textbf{1)} the task is inferred and \textbf{2)} the corresponding mask is used to obtain outputs $\p$. The class probabilities $\p$ correspond to the classes for the inferred task, effectively reusing the neurons in the final layer. 

\ac could use an output size of $\ell$, though we find in practice that it helps significantly to add extra neurons to the final layer. Specifically we consider outputs $\p \in \R^{s}$ and refer to the neurons $\{\ell+1,...,s\}$ as superfluous neurons (s-neurons). The standard cross-entropy loss will push the values of s-neurons down throughout training. Accordingly, we consider an objective $\Gcal$ which encourages the s-neurons to have large negative values and can be used as an alternative to entropy in Equation~\ref{eq:oneshot}. Given data from task $j$, mask $M^j$ will minimize the values of the s-neurons as it was trained to do. Other masks were also trained to minimize the values of the s-neurons, but not for data from task $j$. In Lemma 1 of Section~\ref{sec:analysis} we provide the exact form of $\Gcal$ in code ($ \Gcal = \texttt{logsumexp}\round{\p}$ with masked gradients for $\p_1,...,\p_\ell$) and offer an alternative perspective on why $\Gcal$ is effective --- the gradient of $\Gcal$ for all s-neurons exactly mirrors the gradient from the supervised training loss.

\section{Experiments} \label{sec:exps}
\begin{figure}[t]
    \centering
    \includegraphics[width=\textwidth]{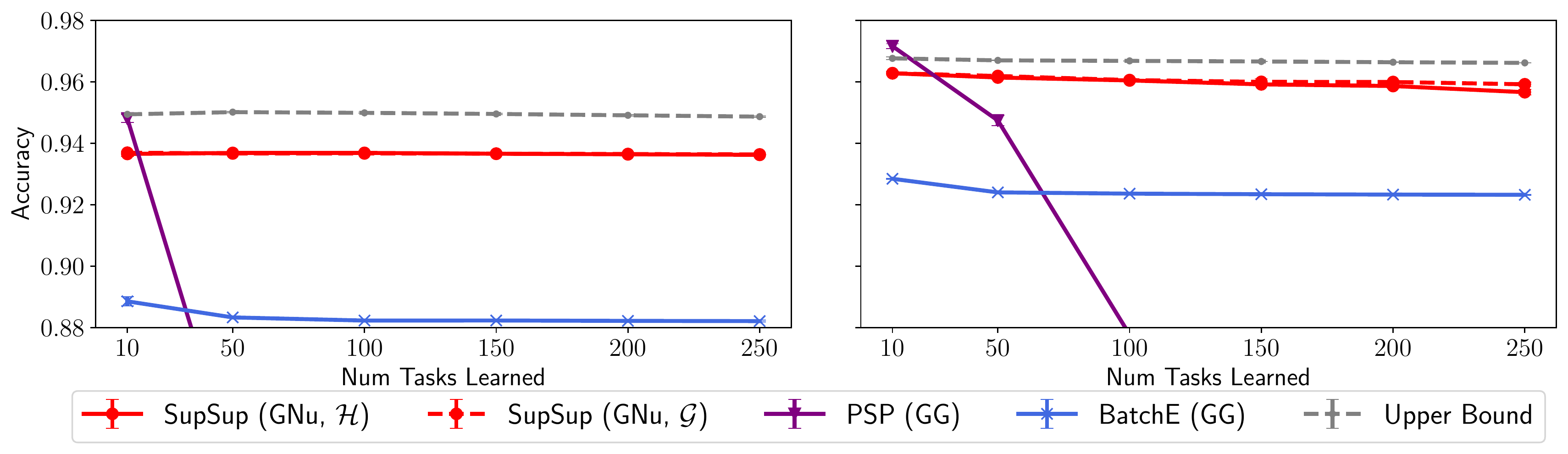}
    \caption{Using \textbf{One-Shot} to infer task identity, \ac outperforms methods with access to task identity. Results shown for PermutedMNIST with LeNet 300-100 \textbf{(left)} and FC 1024-1024 \textbf{(right)}.}
    \label{fig:v1-v2}
    \vspace{-.2em}
\end{figure}

\subsection{Scenario \casename{GG}: Task Identity Information Given During Train and Inference}\label{sec:exp-s1}

\textbf{Datasets, Models \& Training } In this experiment we validate the performance of \ac on SplitCIFAR100 and SplitImageNet. Following Wen \textit{et al.} \cite{wen2020batchensemble}, SplitCIFAR100 randomly partitions CIFAR100 \cite{cifar} into 20 different 5-way classification problems. 
Similarly, SplitImageNet randomly splits the ImageNet \cite{imagenet} dataset into 100 different 10-way classification tasks.
Following \cite{wen2020batchensemble} we use a ResNet-18 with fewer channels for SplitCIFAR100 and a standard ResNet-50 \cite{he2016deep} for SplitImageNet. The Edge-Popup algorithm from \cite{ramanujan2019s} is used to obtain supermasks for various sparsities with a layer-wise budget from  \cite{mocanu2018scalable}. We either initialize each new mask randomly (as in \cite{ramanujan2019s}) or use a running mean of all previous learned masks. This simple method of ``Transfer'' works very well, as illustrated by \autoref{fig:t1}. Additional training details and hyperparameters are provided in Section~\ref{sec:hyperparams}.

\textbf{Computation } In Scenario \casename{GG}, the primary advantage of \ac from Mallya \textit{et al.} \cite{mallya2018packnet} or Wen \textit{et al.} \cite{wen2020batchensemble} is that \ac does not require the base model $W$ to be stored. Since $W$ is random it suffices to store only the random seed. For a fair comparison we also train BatchE \cite{wen2020batchensemble} with random weights. The sparse supermasks are stored in the standard \texttt{scipy.sparse.csc}\footnote{\small{\url{https://docs.scipy.org/doc/scipy/reference/sparse.html}}} format with 16 bit integers. Moreover, \ac requires minimal overhead in terms of forwards pass compute. Elementwise product by a binary mask can be implemented via memory access, \textit{i.e.} selecting indices. Modern GPUs have very high memory bandwidth so the time cost of this operation is small with respect to the time of a forward pass. In particular, on a 1080 Ti this operation requires $\sim 1\%$ of the forward pass time for a ResNet-50, less than the overhead of BatchE (computation in Section~\ref{sec:hyperparams}).
\begin{figure}[t]
    \centering
    \includegraphics[width=0.94\textwidth]{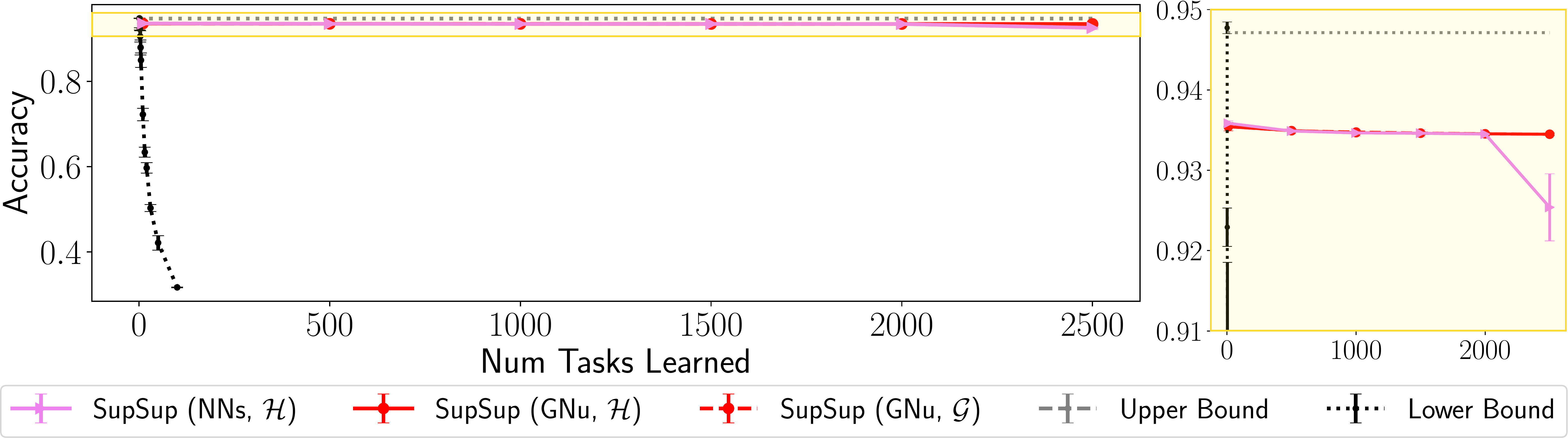}
    \caption{Learning 2500 tasks and inferring task identity using the \textbf{One-Shot} algorithm. Results for both the \casename{GNu} and \casename{NNs} scenarios with the LeNet 300-100 model using output size 500.}
    \label{fig:long-zoom}
    \vspace{-1em}
\end{figure}

\textbf{Baselines } In \autoref{fig:t1}, for ``Separate Heads'' we train different heads for each task using a \emph{trunk} (all layers except the final layer) trained on the first task. In contrast ``Separate Heads - Rand W'' uses a random trunk. BatchE results are given with the trunk trained on the first task (as in \cite{wen2020batchensemble}) and random weights $W$. For ``Upper Bound'', individual models are trained for each task. Furthermore, the trunk for task $i$ is trained on tasks $1,...,i$. For ``Lower Bound'' a shared trunk of the network is trained continuously and a separate head is trained for each task. Since catastrophic forgetting occurs we omit ``Lower Bound'' from \autoref{fig:t1} (the SplitCIFAR100 accuracy is 24.5\%).

\subsection{Scenarios \casename{GNs} \& \casename{GNu}: Task Identity Information Given During Train Only} \label{sec:exp-s23}
Our solutions for \casename{GNs} and \casename{GNu} are very similar. Because \casename{GNu} is strictly more difficult, we focus on only evaluating in Scenario \casename{GNu}. For relevant figures we provide a corresponding table in Section~\ref{sec:tables}.

\textbf{Datasets } Experiments are conducted on PermutedMNIST, RotatedMNIST, and SplitMNIST. For PermutedMNIST \cite{kirkpatrick2017overcoming}, new tasks are created with a fixed random permutation of the pixels of MNIST. For RotatedMNIST, images are rotated by 10 degrees to form a new task with 36 tasks in total (similar to \cite{cheung2019superposition}). Finally SplitMNIST partitions MNIST into 5 different 2-way classification tasks, each containing consecutive classes from the original dataset.

\textbf{Training } We consider two architectures: \textbf{1)} a fully connected network with two hidden layers of size 1024 (denoted FC 1024-1024 and used in \cite{cheung2019superposition}) \textbf{2)} the LeNet 300-100 architecture \cite{lecun1989backpropagation} as used in \cite{frankle2018lottery, dettmers2019sparse}. 
For each task we train for 1000 batches of size 128 using the RMSProp optimizer \cite{tieleman2012lecture} with learning rate $0.0001$ which follows the hyperparameters of \cite{cheung2019superposition}. Supermasks are found using the algorithm of Mallya \textit{et al.} \cite{mallya2018packnet} with threshold value 0. However, we initialize the real valued ``scores'' with Kaiming uniform as in \cite{ramanujan2019s}. Training the mask is not a focus of this work, we choose this method as it is fast and we are not concerned about controlling mask sparsity as in Section~\ref{sec:exp-s1}.

\textbf{Evaluation } At test time we perform inference of task identity once for each batch. If task is not inferred correctly then accuracy is 0 for the batch. Unless noted otherwise we showcase results for the most challenging scenario --- when the task identity is inferred using a single image. We use ``Full Batch'' to indicate that all 128 images are used to infer task identity. Moreover, we experiment with both the the entropy $\Hcal$ and $\Gcal$ (Section~\ref{sec:s-neuron}) objectives to perform task identity inference.

\begin{figure}[t]
    \centering
    \includegraphics[width=\textwidth]{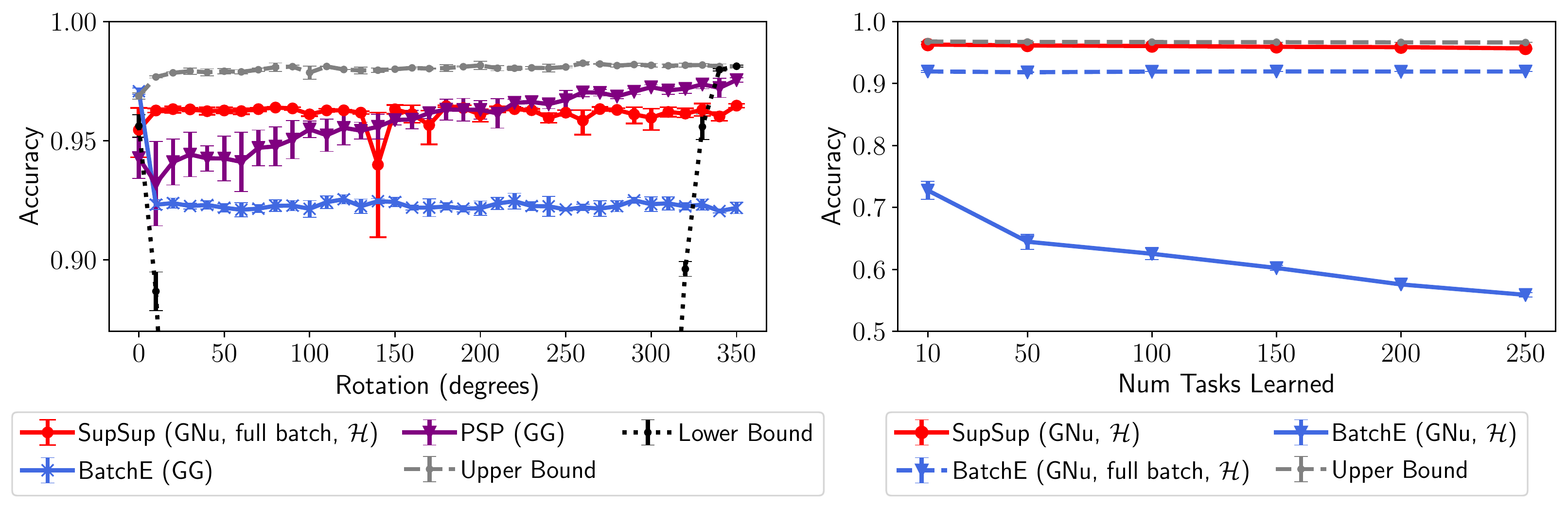}
    \caption{\textbf{(left)} Testing the FC 1024-1024 model on RotatedMNIST. \ac uses \textbf{Binary} to infer task identity with a full batch as tasks are similar (differing by only 10 degrees). \textbf{(right)} The \textbf{One-Shot} algorithm can be used to infer task identity for BatchE \cite{wen2020batchensemble}. Experiment conducted with FC 1024-1024 on PermutedMNIST using an output size of 500, shown as mean and stddev over 3 runs.}
    \label{fig:rot-adapt}
    \vspace{-1em}
\end{figure}

\textbf{Results } \autoref{fig:long-zoom} illustrates that \ac is able to sequentially learn 2500 permutations of MNIST---\ac succeeds in performing 25,000-way classification. This experiment is conducted with the \textbf{One-Shot} algorithm (requiring one gradient computation) using single images to infer task identity. The same trends hold in \autoref{fig:v1-v2}, where \ac outperforms methods which operate in Scenario \casename{GG} by using the \textbf{One-Shot} algorithm to infer task identity. In \autoref{fig:v1-v2}, output sizes of 100 and 500 are respectively used for LeNet 300-100 and FC 1024-1024. The left hand side of \autoref{fig:rot-adapt} illustrates that \ac is able to infer task identity even when tasks are similar---\ac is able to distinguish between rotations of 10 degrees. Since this is a more challenging problem, we use a full batch and the \textbf{Binary} algorithm to perform task identity inference. Figure~\ref{fig:hop-infer} (appendix) shows that for HopSupSup on SplitMNIST, the new mask $\m$ converges to the correct supermask in $<30$ gradient steps.

\textbf{Baselines \& Ablations } \autoref{fig:rot-adapt} (left) shows that even in Scenario \casename{GNu}, \ac is able to outperform PSP~\cite{cheung2019superposition} and BatchE~\cite{wen2020batchensemble} in Scenario \casename{GG}---methods using task identity. We compare SupSup in \casename{GNu} with methods in this strictly easier scenario as they are more competitive. For instance, \cite{van2019three} considers sequential learning problems with only 5-10 tasks. SupSup, after sequentially learning 250 permutations of MNIST, outperforms all non-replay methods from [3] in the \casename{GNu} scenario after they have learned only 10 permutations of MNIST with a similar network. In \casename{GNu}, Online EWC achieves 33.88\% \& SI achieves 29.31\% on 10 permutations of MNIST \cite{van2019three} while SupSup achieves 94.91\% accuracy after 250 permutations (see Table 5 in \cite{van2019three} vs. Table~\ref{tab:rot-adapt-right}). 

\begin{figure}[t]
    \centering
    \includegraphics[width=\textwidth]{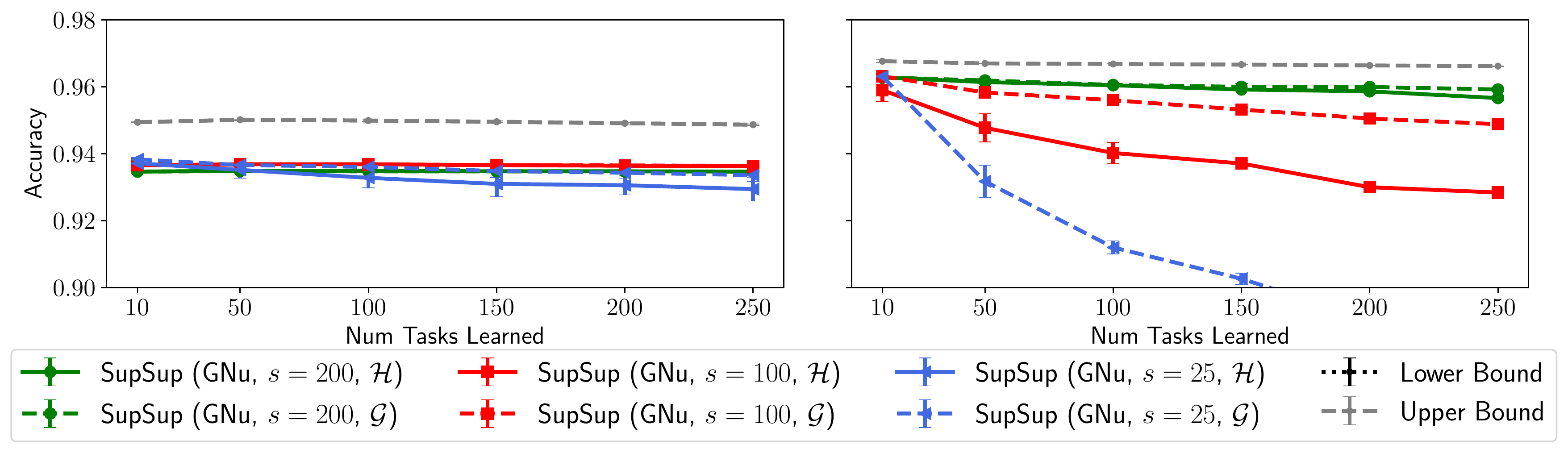}
    \caption{The effect of output size $s$ on \ac performance using the \textbf{One-Shot} algorithm. Results shown for PermutedMNIST with LeNet 300-100 \textbf{(left)} and FC 1024-1024 \textbf{(right)}.}
    \label{fig:v1-v2-sz}
    \vspace{-1.5em}
\end{figure}

In \autoref{fig:rot-adapt} (right) we equip BatchE with task inference using our \textbf{One-Shot} algorithm. Instead of attaching a weight $\alpha_i$ to each supermask, we attach a weight $\alpha_i$ to each rank-one matrix \cite{wen2020batchensemble}. Moreover, in Section~\ref{sec:better-baseline} of the appendix we augment BatchE to perform task-inference using large batch sizes.
``Upper Bound'' and ``Lower Bound'' are the same as in Section~\ref{sec:exp-s1}. Moreover, \autoref{fig:v1-v2-sz} illustrates the importance of output size. Further investigation of this phenomena is provided by Section~\ref{sec:s-neuron} and Lemma 1 of Section~\ref{sec:analysis}. 

\subsection{Scenario \casename{NNs}: No Task Identity During Training or Inference}
\label{sec:exp-s4}
For the \casename{NNs} Scenario we consider PermutedMNIST and train on each task for 1000 batches (the model does not have access to this iteration number). Every 100 batches the model must choose to allocate a new mask or pick an existing mask using the criteria from Section~\ref{sec:S4} ($\epsilon = 2^{-3}$). \autoref{fig:long-zoom} illustrates that without access to any task identity (even during training) SupSup is able to learn thousands of tasks. However, a final dip is observed as a budget of 2500 supermasks total is enforced. \rosanne{TODO: write more about this. Most impressive result!}

\section{Conclusion}
Supermasks in Superposition (\ac) is a flexible and compelling model applicable to a wide range of scenarios in Continual Learning. \ac leverages the power of subnetworks~\cite{zhou2019deconstructing, ramanujan2019s, mallya2018packnet}, and gradient-based optimization to infer task identity when unknown. 
\ac achieves state-of-the-art performance on SplitImageNet when given task identity, and performs well on thousands of permutations and almost indiscernible rotations of MNIST without any task information. 

We observe limitations in applying SupSup with task identity inference to non-uniform and more challenging problems. Task inference fails when models are not well calibrated---are overly confident for the wrong task. As future work, we hope to explore
automatic task inference with more calibrated models~\cite{guo2017calibration}, as well as circumventing calibration challenges by using  optimization objectives such as self-supervision \cite{he2019momentum} and energy based models \cite{grathwohl2019your}. In doing so, we hope to tackle large-scale problems in Scenarios \casename{GN} and \casename{NNs}.

\section*{Broader Impact}

A goal of continual learning is to solve many tasks with a single model. However, it is not exactly clear what qualifies as a \textit{single model}. Therefore, a concrete objective has become to learn many tasks as efficiently as possible. We believe that \ac is a useful step in this direction. However, there are consequences to more efficient models, both positive and negative.

We begin with the positive consequences:
\begin{itemize}
    \item Efficient models require less compute, and are therefore less harmful for the environment then learning one model per task \cite{schwartz2019green}. This is especially true if models are able to leverage information from past tasks, and training on new tasks is then faster.
    \item Efficient models may be run on the end device. This helps to preserve privacy as a user's data does not have to be sent to the cloud for computation.
    \item If models are more efficient then large scale research is not limited to wealthier institutions. These institutions are more likely in privileged parts of the world and may be ignorant of problems facing developing nations. Moreover, privileged institutions may not be a representative sample of the research community.
\end{itemize}
We would also like to highlight and discuss the negative consequences of models which can efficiently learn many tasks, and efficient models in general. When models are more efficient, they are also more available and less subject to regularization and study as a result. For instance, when a high-impact model is released by an institution it will hopefully be accompanied by a Model Card \cite{mitchell2019model} analyzing the bias and intended use of the model. By contrast, if anyone is able to train a powerful model this may no longer be the case, resulting in a proliferation of models with harmful biases or intended use. Taking the United States for instance, bias can be harmful as models show disproportionately more errors for already marginalized groups \cite{buolamwini2018gender}, furthering existing and deeply rooted structural racism.

\subsubsection*{Acknowledgments}
We thank Gabriel Ilharco Magalhães and Sarah Pratt for helpful comments. For valuable conversations we also thank Tim Dettmers, Kiana Ehsani, Ana Marasović, Suchin Gururangan, Zoe Steine-Hanson, Connor Shorten, Samir Yitzhak Gadre, Samuel McKinney and Kishanee Haththotuwegama. This work is in part supported by NSF IIS 1652052, IIS 17303166, DARPA N66001-19-2-4031, DARPA W911NF-15-1-0543 and gifts from Allen Institute for Artificial Intelligence. Additional revenues: co-authors had employment with the Allen Institute for AI.

{
\bibliographystyle{plain}
\bibliography{ref}

\begin{thebibliography}{10}

\bibitem{bengio2006convex}
Yoshua Bengio, Nicolas~L Roux, Pascal Vincent, Olivier Delalleau, and Patrice
  Marcotte.
\newblock Convex neural networks.
\newblock In {\em Advances in neural information processing systems}, pages
  123--130, 2006.

\bibitem{buolamwini2018gender}
Joy Buolamwini and Timnit Gebru.
\newblock Gender shades: Intersectional accuracy disparities in commercial
  gender classification.
\newblock In {\em Conference on fairness, accountability and transparency},
  pages 77--91, 2018.

\bibitem{chaudhry2018efficient}
Arslan Chaudhry, Marc'Aurelio Ranzato, Marcus Rohrbach, and Mohamed Elhoseiny.
\newblock Efficient lifelong learning with a-gem.
\newblock {\em arXiv preprint arXiv:1812.00420}, 2018.

\bibitem{cheung2019superposition}
Brian Cheung, Alexander Terekhov, Yubei Chen, Pulkit Agrawal, and Bruno
  Olshausen.
\newblock Superposition of many models into one.
\newblock In {\em Advances in Neural Information Processing Systems}, pages
  10867--10876, 2019.

\bibitem{imagenet}
Jia Deng, Wei Dong, Richard Socher, Li-Jia Li, Kai Li, and Li~Fei-Fei.
\newblock Imagenet: A large-scale hierarchical image database.
\newblock In {\em CVPR 2009}, 2009.

\bibitem{dettmers2019sparse}
Tim Dettmers and Luke Zettlemoyer.
\newblock Sparse networks from scratch: Faster training without losing
  performance.
\newblock {\em arXiv preprint arXiv:1907.04840}, 2019.

\bibitem{frank1956algorithm}
Marguerite Frank and Philip Wolfe.
\newblock An algorithm for quadratic programming.
\newblock {\em Naval research logistics quarterly}, 3(1-2):95--110, 1956.

\bibitem{frankle2018lottery}
Jonathan Frankle and Michael Carbin.
\newblock The lottery ticket hypothesis: Finding sparse, trainable neural
  networks.
\newblock {\em arXiv preprint arXiv:1803.03635}, 2018.

\bibitem{frankle2020training}
Jonathan Frankle, David~J Schwab, and Ari~S Morcos.
\newblock Training batchnorm and only batchnorm: On the expressive power of
  random features in cnns.
\newblock {\em arXiv preprint arXiv:2003.00152}, 2020.

\bibitem{french1999catastrophic}
Robert~M French.
\newblock Catastrophic forgetting in connectionist networks.
\newblock {\em Trends in cognitive sciences}, 3(4):128--135, 1999.

\bibitem{golkar2019continual}
Siavash Golkar, Michael Kagan, and Kyunghyun Cho.
\newblock Continual learning via neural pruning.
\newblock {\em arXiv preprint arXiv:1903.04476}, 2019.

\bibitem{goodfellow2013empirical}
Ian~J Goodfellow, Mehdi Mirza, Da~Xiao, Aaron Courville, and Yoshua Bengio.
\newblock An empirical investigation of catastrophic forgetting in
  gradient-based neural networks.
\newblock {\em arXiv preprint arXiv:1312.6211}, 2013.

\bibitem{grathwohl2019your}
Will Grathwohl, Kuan-Chieh Wang, J{\"o}rn-Henrik Jacobsen, David Duvenaud,
  Mohammad Norouzi, and Kevin Swersky.
\newblock Your classifier is secretly an energy based model and you should
  treat it like one.
\newblock {\em arXiv preprint arXiv:1912.03263}, 2019.

\bibitem{guo2017calibration}
Chuan Guo, Geoff Pleiss, Yu~Sun, and Kilian~Q Weinberger.
\newblock On calibration of modern neural networks.
\newblock In {\em Proceedings of the 34th International Conference on Machine
  Learning-Volume 70}, pages 1321--1330. JMLR. org, 2017.

\bibitem{ha2016hypernetworks}
David Ha, Andrew Dai, and Quoc~V Le.
\newblock Hypernetworks.
\newblock {\em arXiv preprint arXiv:1609.09106}, 2016.

\bibitem{he2019momentum}
Kaiming He, Haoqi Fan, Yuxin Wu, Saining Xie, and Ross Girshick.
\newblock Momentum contrast for unsupervised visual representation learning.
\newblock {\em arXiv preprint arXiv:1911.05722}, 2019.

\bibitem{he2015delving}
Kaiming He, Xiangyu Zhang, Shaoqing Ren, and Jian Sun.
\newblock Delving deep into rectifiers: Surpassing human-level performance on
  imagenet classification.
\newblock In {\em Proceedings of the IEEE international conference on computer
  vision}, pages 1026--1034, 2015.

\bibitem{he2016deep}
Kaiming He, Xiangyu Zhang, Shaoqing Ren, and Jian Sun.
\newblock Deep residual learning for image recognition.
\newblock In {\em Proceedings of the IEEE conference on computer vision and
  pattern recognition}, pages 770--778, 2016.

\bibitem{hendrycks2016baseline}
Dan Hendrycks and Kevin Gimpel.
\newblock A baseline for detecting misclassified and out-of-distribution
  examples in neural networks.
\newblock {\em arXiv preprint arXiv:1610.02136}, 2016.

\bibitem{hopfield1982neural}
John~J Hopfield.
\newblock Neural networks and physical systems with emergent collective
  computational abilities.
\newblock {\em Proceedings of the national academy of sciences},
  79(8):2554--2558, 1982.

\bibitem{ioffe2015batch}
Sergey Ioffe and Christian Szegedy.
\newblock Batch normalization: Accelerating deep network training by reducing
  internal covariate shift.
\newblock {\em arXiv preprint arXiv:1502.03167}, 2015.

\bibitem{kingma2014adam}
Diederik~P Kingma and Jimmy Ba.
\newblock Adam: A method for stochastic optimization.
\newblock {\em arXiv preprint arXiv:1412.6980}, 2014.

\bibitem{kirkpatrick2017overcoming}
James Kirkpatrick, Razvan Pascanu, Neil Rabinowitz, Joel Veness, Guillaume
  Desjardins, Andrei~A Rusu, Kieran Milan, John Quan, Tiago Ramalho, Agnieszka
  Grabska-Barwinska, et~al.
\newblock Overcoming catastrophic forgetting in neural networks.
\newblock {\em Proceedings of the national academy of sciences},
  114(13):3521--3526, 2017.

\bibitem{cifar}
Alex Krizhevsky.
\newblock Learning multiple layers of features from tiny images.
\newblock Technical report, University of Toronto, 2009.

\bibitem{lecun1989backpropagation}
Yann LeCun, Bernhard Boser, John~S Denker, Donnie Henderson, Richard~E Howard,
  Wayne Hubbard, and Lawrence~D Jackel.
\newblock Backpropagation applied to handwritten zip code recognition.
\newblock {\em Neural computation}, 1(4):541--551, 1989.

\bibitem{mnist}
Yann LeCun, Corinna Cortes, and CJ~Burges.
\newblock Mnist handwritten digit database.
\newblock 2010.

\bibitem{lopez2017gradient}
David Lopez-Paz and Marc'Aurelio Ranzato.
\newblock Gradient episodic memory for continual learning.
\newblock In {\em Advances in Neural Information Processing Systems}, pages
  6467--6476, 2017.

\bibitem{cosine}
Ilya Loshchilov and Frank Hutter.
\newblock Sgdr: Stochastic gradient descent with warm restarts, 2016.

\bibitem{malach2020proving}
Eran Malach, Gilad Yehudai, Shai Shalev-Shwartz, and Ohad Shamir.
\newblock Proving the lottery ticket hypothesis: Pruning is all you need.
\newblock {\em arXiv preprint arXiv:2002.00585}, 2020.

\bibitem{mallya2018piggyback}
Arun Mallya, Dillon Davis, and Svetlana Lazebnik.
\newblock Piggyback: Adapting a single network to multiple tasks by learning to
  mask weights.
\newblock In {\em Proceedings of the European Conference on Computer Vision
  (ECCV)}, pages 67--82, 2018.

\bibitem{mallya2018packnet}
Arun Mallya and Svetlana Lazebnik.
\newblock Packnet: Adding multiple tasks to a single network by iterative
  pruning.
\newblock In {\em Proceedings of the IEEE Conference on Computer Vision and
  Pattern Recognition}, pages 7765--7773, 2018.

\bibitem{masse2018alleviating}
Nicolas~Y Masse, Gregory~D Grant, and David~J Freedman.
\newblock Alleviating catastrophic forgetting using context-dependent gating
  and synaptic stabilization.
\newblock {\em Proceedings of the National Academy of Sciences},
  115(44):E10467--E10475, 2018.

\bibitem{mccloskey1989catastrophic}
Michael McCloskey and Neal~J Cohen.
\newblock Catastrophic interference in connectionist networks: The sequential
  learning problem.
\newblock In {\em Psychology of learning and motivation}, volume~24, pages
  109--165. Elsevier, 1989.

\bibitem{mitchell2019model}
Margaret Mitchell, Simone Wu, Andrew Zaldivar, Parker Barnes, Lucy Vasserman,
  Ben Hutchinson, Elena Spitzer, Inioluwa~Deborah Raji, and Timnit Gebru.
\newblock Model cards for model reporting.
\newblock In {\em Proceedings of the conference on fairness, accountability,
  and transparency}, pages 220--229, 2019.

\bibitem{mocanu2018scalable}
Decebal~Constantin Mocanu, Elena Mocanu, Peter Stone, Phuong~H Nguyen,
  Madeleine Gibescu, and Antonio Liotta.
\newblock Scalable training of artificial neural networks with adaptive sparse
  connectivity inspired by network science.
\newblock {\em Nature communications}, 9(1):1--12, 2018.

\bibitem{pascanu2013revisiting}
Razvan Pascanu and Yoshua Bengio.
\newblock Revisiting natural gradient for deep networks.
\newblock {\em arXiv preprint arXiv:1301.3584}, 2013.

\bibitem{paszke2019pytorch}
Adam Paszke, Sam Gross, Francisco Massa, Adam Lerer, James Bradbury, Gregory
  Chanan, Trevor Killeen, Zeming Lin, Natalia Gimelshein, Luca Antiga, et~al.
\newblock Pytorch: An imperative style, high-performance deep learning library.
\newblock In {\em Advances in Neural Information Processing Systems}, pages
  8024--8035, 2019.

\bibitem{ramachandran2017searching}
Prajit Ramachandran, Barret Zoph, and Quoc~V Le.
\newblock Searching for activation functions.
\newblock {\em arXiv preprint arXiv:1710.05941}, 2017.

\bibitem{ramanujan2019s}
Vivek Ramanujan, Mitchell Wortsman, Aniruddha Kembhavi, Ali Farhadi, and
  Mohammad Rastegari.
\newblock What's hidden in a randomly weighted neural network?
\newblock {\em arXiv preprint arXiv:1911.13299}, 2019.

\bibitem{rebuffi2017icarl}
Sylvestre-Alvise Rebuffi, Alexander Kolesnikov, Georg Sperl, and Christoph~H
  Lampert.
\newblock icarl: Incremental classifier and representation learning.
\newblock In {\em Proceedings of the IEEE conference on Computer Vision and
  Pattern Recognition}, pages 2001--2010, 2017.

\bibitem{rolnick2019experience}
David Rolnick, Arun Ahuja, Jonathan Schwarz, Timothy Lillicrap, and Gregory
  Wayne.
\newblock Experience replay for continual learning.
\newblock In {\em Advances in Neural Information Processing Systems}, pages
  348--358, 2019.

\bibitem{rusu2016progressive}
Andrei~A Rusu, Neil~C Rabinowitz, Guillaume Desjardins, Hubert Soyer, James
  Kirkpatrick, Koray Kavukcuoglu, Razvan Pascanu, and Raia Hadsell.
\newblock Progressive neural networks.
\newblock {\em arXiv preprint arXiv:1606.04671}, 2016.

\bibitem{schrauwen2007overview}
Benjamin Schrauwen, David Verstraeten, and Jan Van~Campenhout.
\newblock An overview of reservoir computing: theory, applications and
  implementations.
\newblock In {\em Proceedings of the 15th european symposium on artificial
  neural networks. p. 471-482 2007}, pages 471--482, 2007.

\bibitem{schwartz2019green}
Roy Schwartz, Jesse Dodge, Noah~A Smith, and Oren Etzioni.
\newblock Green ai. corr abs/1907.10597 (2019).
\newblock {\em arXiv preprint arXiv:1907.10597}, 2019.

\bibitem{shin2017continual}
Hanul Shin, Jung~Kwon Lee, Jaehong Kim, and Jiwon Kim.
\newblock Continual learning with deep generative replay.
\newblock In {\em Advances in Neural Information Processing Systems}, pages
  2990--2999, 2017.

\bibitem{storkey1997increasing}
Amos Storkey.
\newblock Increasing the capacity of a hopfield network without sacrificing
  functionality.
\newblock In {\em International Conference on Artificial Neural Networks},
  pages 451--456. Springer, 1997.

\bibitem{thrun1998lifelong}
Sebastian Thrun.
\newblock Lifelong learning algorithms.
\newblock In {\em Learning to learn}, pages 181--209. Springer, 1998.

\bibitem{tieleman2012lecture}
Tijmen Tieleman and Geoffrey Hinton.
\newblock Lecture 6.5-rmsprop: Divide the gradient by a running average of its
  recent magnitude.
\newblock {\em COURSERA: Neural networks for machine learning}, 4(2):26--31,
  2012.

\bibitem{van2019three}
Gido~M van~de Ven and Andreas~S Tolias.
\newblock Three scenarios for continual learning.
\newblock {\em arXiv preprint arXiv:1904.07734}, 2019.

\bibitem{Oswald2020Continual}
Johannes von Oswald, Christian Henning, João Sacramento, and Benjamin~F.
  Grewe.
\newblock Continual learning with hypernetworks.
\newblock In {\em International Conference on Learning Representations}, 2020.

\bibitem{wen2020batchensemble}
Yeming Wen, Dustin Tran, and Jimmy Ba.
\newblock Batchensemble: an alternative approach to efficient ensemble and
  lifelong learning.
\newblock {\em arXiv preprint arXiv:2002.06715}, 2020.

\bibitem{xu2018reinforced}
Ju~Xu and Zhanxing Zhu.
\newblock Reinforced continual learning.
\newblock In {\em Advances in Neural Information Processing Systems}, pages
  899--908, 2018.

\bibitem{yoon2017lifelong}
Jaehong Yoon, Eunho Yang, Jeongtae Lee, and Sung~Ju Hwang.
\newblock Lifelong learning with dynamically expandable networks.
\newblock {\em arXiv preprint arXiv:1708.01547}, 2017.

\bibitem{zenke2017continual}
Friedemann Zenke, Ben Poole, and Surya Ganguli.
\newblock Continual learning through synaptic intelligence.
\newblock In {\em Proceedings of the 34th International Conference on Machine
  Learning-Volume 70}, pages 3987--3995. JMLR. org, 2017.

\bibitem{zeno2018task}
Chen Zeno, Itay Golan, Elad Hoffer, and Daniel Soudry.
\newblock Task agnostic continual learning using online variational bayes.
\newblock {\em arXiv preprint arXiv:1803.10123}, 2018.

\bibitem{zhao1996incremental}
Jieyu Zhao and Jurgen Schmidhuber.
\newblock Incremental self-improvement for life-time multi-agent reinforcement
  learning.
\newblock In {\em From Animals to Animats 4: Proceedings of the Fourth
  International Conference on Simulation of Adaptive Behavior, Cambridge, MA},
  pages 516--525, 1996.

\bibitem{zhou2019deconstructing}
Hattie Zhou, Janice Lan, Rosanne Liu, and Jason Yosinski.
\newblock Deconstructing lottery tickets: Zeros, signs, and the supermask.
\newblock In {\em Advances in Neural Information Processing Systems}, pages
  3592--3602, 2019.

\end{thebibliography}
}

\clearpage

\appendix

\section{Algorithm pseudo-code} \label{sec:pseudo}
Algorithms \ref{alg:oneshot} and \ref{alg:binary} respectively provide pseudo-code for the \textbf{One-Shot} and \textbf{Binary} algorithms detailed in Section~\ref{sec:S23}. Both aim to infer the task $j \in \{1,...,k\}$ associated with input data $\x$ by minimizing the objective $\Hcal$.

\begin{algorithm}[t]
\caption{One-Shot$(f, \x, W, k, \{M^i\}_{i=1}^k, \Hcal)$}\label{alg:oneshot}
\begin{algorithmic}[1]
\State{$\alpha \gets \begin{bmatrix} \frac{1}{k} & \frac{1}{k} &... &\frac{1}{k}  \end{bmatrix}$} {\color{gray}\Comment{Initialize $\alpha$}}
\State{$\p \gets f\round{\x, W \odot \round{ \sum_{i=1}^k \alpha_i M^i}} $} {\color{gray}\Comment{Superimposed output}}
\State{$\textbf{return} \ \argmax_i  \round{- \frac{ \partial \Hcal\round{\p}}{\partial \alpha_i}}$} {\color{gray}\Comment{Return coordinate for which objective maximally decreasing}}
\end{algorithmic}
\end{algorithm}

\begin{algorithm}[t] 
\caption{Binary$(f, \x, W, k, \{M^i\}_{i=1}^k, \Hcal)$}\label{alg:binary}
\begin{algorithmic}[1] 
\State{$\alpha \gets \begin{bmatrix} \frac{1}{k} & \frac{1}{k} &... &\frac{1}{k}  \end{bmatrix}$} {\color{gray}\Comment{Initialize $\alpha$}}
\While{$\norm{0}{\alpha} > 1$} {\color{gray}\Comment{Iterate until $\alpha$ has a single nonzero entry}}
    \State{$\p \gets f\round{\x, W \odot \round{ \sum_{i=1}^k \alpha_i M^i}} $} {\color{gray}\Comment{Superimposed output}}
    \State{$g \gets - \nabla_\alpha \Hcal\round{\p}$} {\color{gray}\Comment{Gradient of objective}}
    \For{$i \in \{1,...,k\}$} {\color{gray}\Comment{In code this \textbf{for} loop is vectorized}}
        \If{$g_i \leq \textbf{median}\round{g}$}
            \State{$\alpha_i \gets 0$} {\color{gray}\Comment{Zero out $\alpha_i$ for which objective minimally decreasing}}
        \EndIf
    \EndFor
    \State{$\alpha \gets \alpha /\norm{1}{\alpha} $} {\color{gray}\Comment{Re-normalize $\alpha$ to sum to 1}}
\EndWhile
\State{$\textbf{return} \ \argmax_i  \alpha_i$}
\end{algorithmic}
\end{algorithm}

\section{Extended Details for Hop\ac} \label{sec:hop-extended}

This section provides further details and experiments for Hop\ac (introduced in Section~\ref{sec:hop}). Hop\ac provides a method for storing the growing set of supermasks in a fixed size reservoir instead of explicitly storing each mask.
\subsection{Training}
Recall that Hop\ac operates in Scenario \casename{GNu} and so task identity is known during training. Instead of explicitly storing each mask, we will instead store two fixed sized variables $\Psi$ and  $\mu$ which are both initially $\zero$. The weights of the Hopfield network are $\Psi$ and $\mu$ stores a running mean of all masks learned so far. For a new task $k$ we use the same algorithm as in Section~\ref{sec:exp-s23} to learn a binary mask $\m^i$ which performs well for task $k$. Since Hopfield networks consider binary strings in $\{-1,1\}^d$ and we use masks $\m^i \in \{0,1\}^d$ we will consider $\z^k = 2\m^k - 1$. In practice we then update $\Psi$ and $\mu$ as 
\begin{align}
    &\Psi \gets \Psi + \frac{1}{d}\round{ \z^k {\z^k}^\top - \z^k \round{\Psi \z^k}^\top - \round{\Psi \z^k} {\z^k}^\top -  \text{Id}},  & \mu \gets \frac{k-1}{k} \mu + \frac{1}{k} \z^k 
\end{align}
where Id is the identity matrix. This update rule for $\Psi$ is referred to as the Storkey learning rule \cite{storkey1997increasing} and is more expressive than the alternative---the Hebbian rule $\Psi \gets \Psi + \frac{1}{d} \z^k {\z^k}^\top$ \cite{hopfield1982neural} provided for brevity in Section~\ref{sec:S23}. With either update rules the learned $\z^i$ will be a minimizer of the Hopfield energy $E_{\Psi}(\z) = \sum_{uv} \Psi_{uv} \z_u \z_v$.
\begin{figure}[t]
    \centering
    \includegraphics[width=\textwidth]{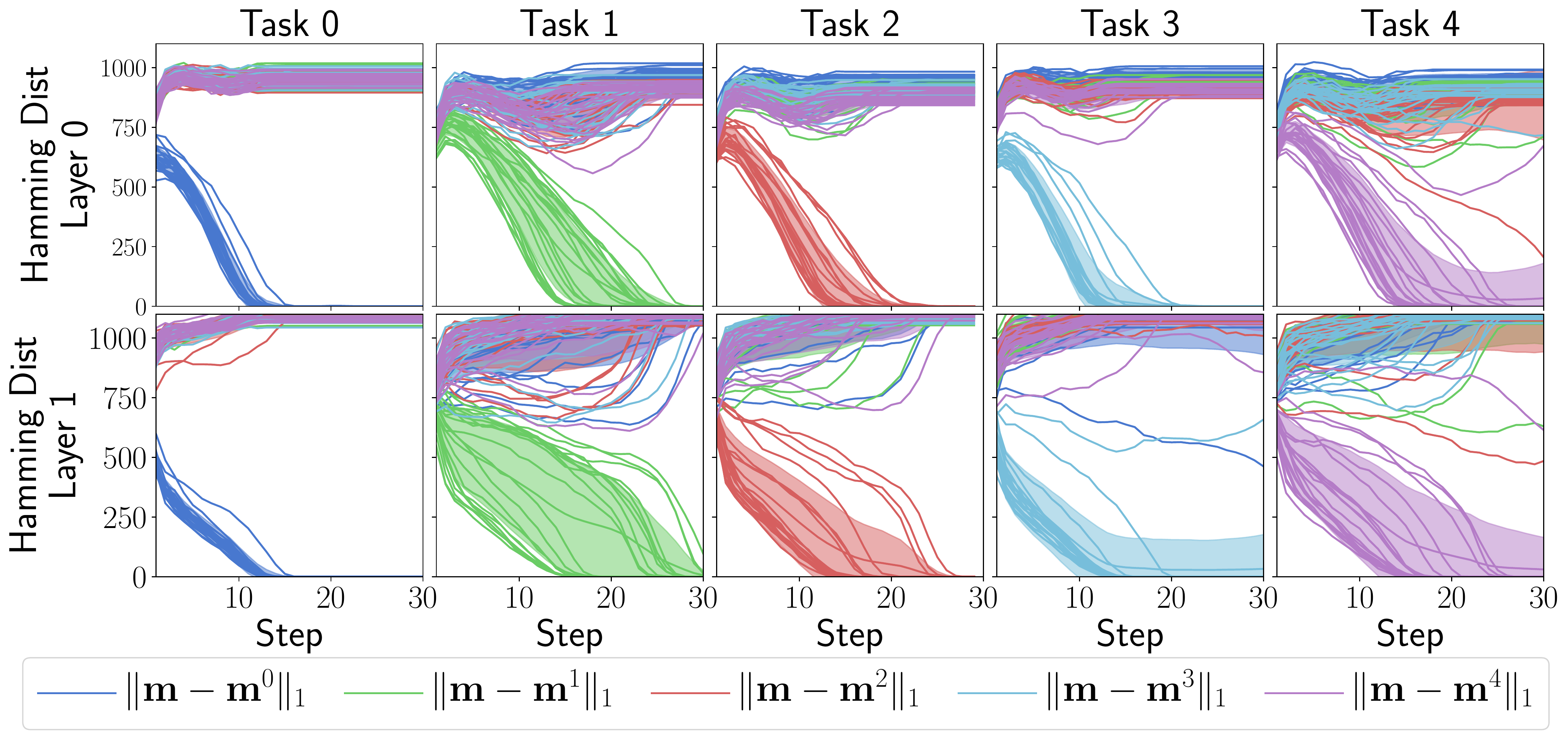}
    \caption{During \emph{Hopfield Recovery} the new mask $\m$ converges to the correct mask learned during training. Note that $\m^i$ denotes the mask learned for task $i$.}
    \label{fig:hop-infer}
\end{figure}
\subsection{Inference}
During inference we receive data $\x$ from some task $j$, but this task information is not given to the model. Hop\ac first initializes a new binary string $\z$ with $\mu$. Next, Hop\ac uses gradient descent to minimize the Hopfield energy in conjunction with the output entropy using mask $\m = \frac{1}{2}\z + 1$, a process we refer to as \emph{Hopfield Recovery}. Minimizing the energy will hopefully push $\m$ (equivalently $\z$) towards a mask learned during training and minimizing the entropy will hopefully push $\m$ towards the correct mask $\m^j$. We may then use the recovered mask to compute the network output.

In practice we use one pass through the evaluation set (with batch size 64, requiring $T \approx 30$ steps) to recover a mask and another to perform evaluation with the recovered mask. When recovering the mask we gradually increase the strength of the Hopfield term and decrease the strength of the entropy term. Otherwise the Hopfield term initially pulls $\z$ in the wrong direction or the final $\z$ does not lie at a minimum of $E_\Psi$. For step $t \in \{1,...,T\}$, and constant $\gamma$ we use the objective $\Jcal$ as
\begin{align}
    \Jcal(\z, t) = \frac{\gamma t}{T} E_{\Psi}(\z) +  \round{1 - \frac{t}{T}}  \mathcal{H}\round{\p}
\end{align}
where $\p$ denotes the output using mask $\m = \frac{1}{2}\z + 1$. 

\begin{figure}
\centering
\begin{minipage}{.6\textwidth}
  \centering
  \includegraphics[width=.8\linewidth]{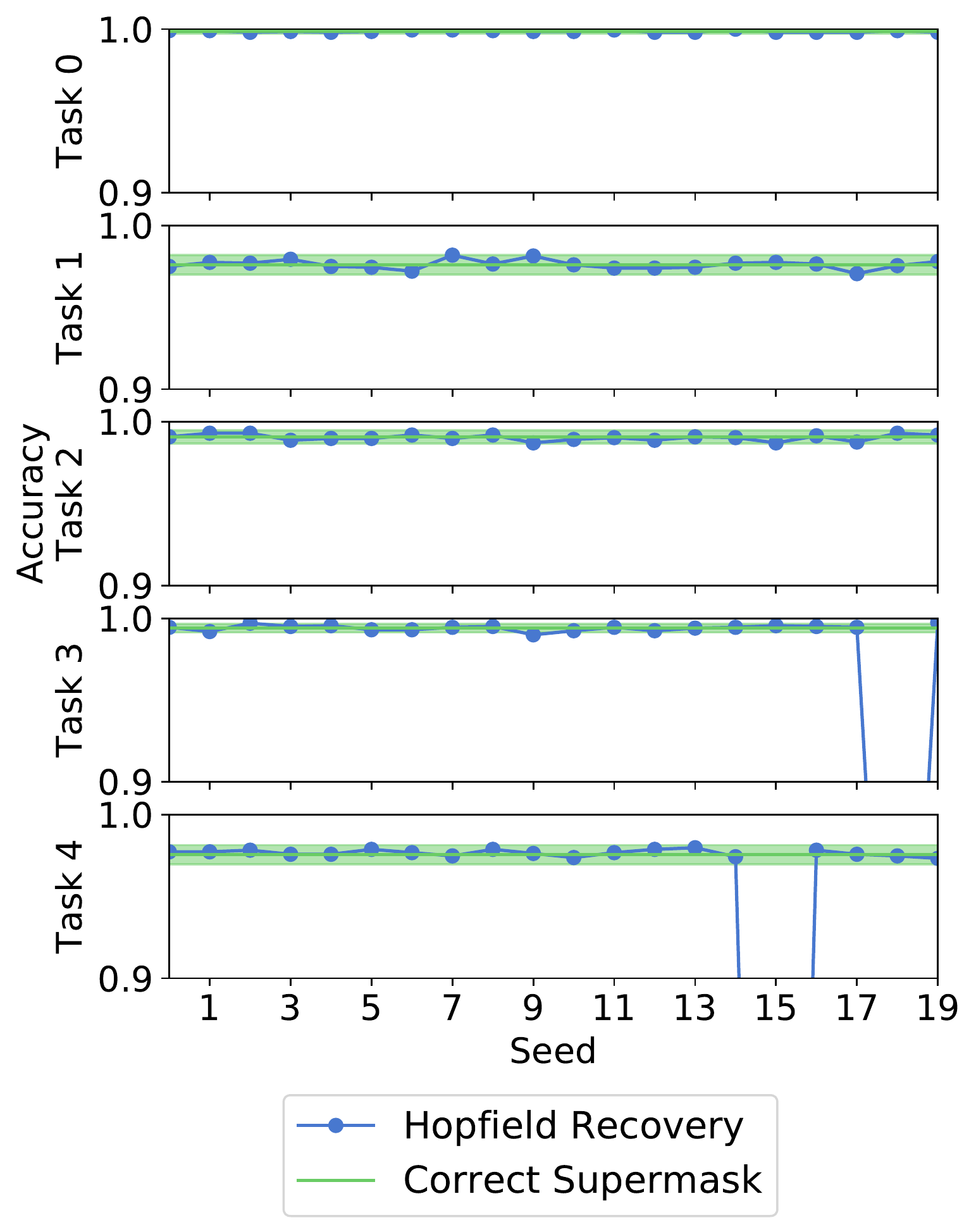}
  \captionof{figure}{Evaluating (with 20 random seeds) on SplitMNIST after finding a mask with \emph{Hopfield Recovery}. Average accuracy is 97.43\%.}
  \label{fig:hop-res}
\end{minipage}%
\begin{minipage}{.4\textwidth}
  \centering
\begin{tikzpicture}
  [
    grow                    = down,
    level 1/.style          = {sibling distance = 6em, level distance=4em},
    level 2/.style          = {sibling distance = 6em, level distance=4em},
    level 3/.style          = {sibling distance = 3.5em, level distance=3.5em},
    level distance          = 4.5em,
    edge from parent/.style = {draw, -latex},
    every node/.style       = {font=\scriptsize},
    >=latex
  ]
  \node (tiidinf) [root] [treenode2] {Task ID given during\\ train \& inference}
    child { node (s1) [env2] {\casename{GG}}
        edge from parent node [left] {Yes}
    }
    child { node (tiidtrain) [root] [treenode2] {Task ID given\\ during training}
        child { node (tsly) [root] [treenode2] {Tasks share\\labels}
            child { node (s2) [root] [env2] {\casename{GNs}}
                child[level distance=2.4em] 
                { node (TL) [whiteenv] {} edge from parent[draw=none]
                }
                edge from parent node [left] {Yes}
            }
            child { node (s3) [root] [env2] {\casename{GNu}}
                edge from parent node [right] {No}
            }
            edge from parent node [left] {Yes}
        }
        child { node (tsln) [root] [treenode2] {Tasks share\\labels}
            child { node (s4) [root] [env2] {\casename{NNs}}
                edge from parent node [right] {Yes}
            }
            edge from parent node [right] {No}
        }
        edge from parent node [right]  {No} 
    };
\end{tikzpicture}
\caption{Continual learning scenarios detailed in Table~\ref{tab:scenarios} represented in a tree graph, as in \cite{zeno2018task}.}
\label{fig:scenarios}
\end{minipage}
\end{figure}

Figure~\ref{fig:hop-infer} illustrates that after approximately 30 steps of gradient descent on $\z$ using objective $\Jcal$, the mask $\m = \frac{1}{2}\z + 1$ converges to the correct mask learned during training. This experiment is conducted for 20 different random seeds on SplitMNIST (see Section~\ref{sec:exp-s23}) training for 1 epoch per task. Evaluation with the recovered mask for each seed is then given by Figure~\ref{fig:hop-res}. As expected, when the correct mask is successfully recovered, accuracy matches directly using the correct mask. For hyperparameters we set $\gamma = 1.5 \cdot 10^{-3}$ and perform gradient descent during Hopfield recovery with learning rate $0.5 \cdot 10^3$, momentum $0.9$, and weight decay $10^{-4}$. 

\subsection{Network Architecture}
Let $\texttt{BN}$ denote non-affine batch normalization \cite{ioffe2015batch}, \textit{i.e.} batch normalization with no learned parameters. Also recall that we are masking layer outputs instead of weights, and the weights still remain fixed (see Section~\ref{sec:hop}). Therefore, with mask $\m = (\m_1, \m_2)$ and weights $W = (W_1, W_2, W_3)$ we compute outputs as
\begin{align}
    f(\x, \m, W) = \texttt{softmax}\round{ W_3^\top \sigma\round{\m_2 \odot \texttt{BN}\round{ W_2^\top \sigma\round{\m_1 \odot \texttt{BN}\round{W_1^\top \x }} }}}
\end{align}
where $\sigma$ denotes the Swish nonlinearity \cite{ramachandran2017searching}. Without masking or normalization $f$ is a fully connected network with two hidden layers of size 2048. We also note that Hop\ac requires 10 output neurons for SplitMNIST in Scenario \casename{GNu}, and the composition of non-affine batch normalization with a binary mask was inspired by BatchNets \cite{frankle2020training}. 

\section{Augmenting BatchE For Scnario \casename{GNu}} \label{sec:better-baseline}

In Section~\ref{sec:exp-s23} we demonstrate that BatchE \cite{wen2020batchensemble} is able to infer task identity using the \textbf{One-Shot} algorithm. In this section we show that, equipped with $\Hcal$ from Section~\ref{sec:method}, BatchE can also infer task identity by using a large batch size. We refer to this method as Augmented BatchE (ABatchE).

For clarity we describe ABatchE for one linear layer, \textit{i.e.} we describe the application of ABatchE to
\begin{align}
    f(\x, W) = \texttt{softmax}\round{W^\top \x}
\end{align}
for input data $\x \in \R^m$ and weights  $W \in \R^{m \times n}$. In BatchE \cite{wen2020batchensemble}, $W$ is trained on the first task then frozen. For task $i$ BatchE learns ``fast weights'' $r_i \in \R^{m}$, $s_i \in \R^{n}$ and outputs are computed via
\begin{align} \label{eq:ogbe}
    f(\x, W) = \texttt{softmax}\round{\round{W \odot r_i s_i^\top  }^\top \x}.
\end{align}
Wen \textit{et al.} \cite{wen2020batchensemble} further demonstrate that Equation~\ref{eq:ogbe} can be vectorized as
\begin{align} \label{eq:vbe}
    f(\x, W) = \texttt{softmax}\round{ \round{W^\top \round{\x \odot r_i }} \odot s_i }
\end{align}
or, for a batch of data $X \in \R^{b \times m}$,
\begin{align} \label{eq:bvbe}
    f(X, W) = \texttt{softmax}\round{ \round{\round{X \odot R^{b}_i }  W} \odot S^{b}_i }.
\end{align}
In Equation~\ref{eq:bvbe}, $R^b_i \in \R^{b \times m}$ is a matrix where each of the $b$ rows is $r_i$ (likewise $S^b_i \in \R^{b \times n}$ is a matrix where each of the $b$ rows is $s_i$).

As in Section~\ref{sec:S23} we now consider the case where data $X \in \R^{b \times m}$ comes from task $j$ but this information is not known to the model. For ABatchE we repeat the data $k$ times, where $k$ is the number of tasks learned so far, and use different ``fast weights'' for each repetiton. Specifically, we consider repeated data $\tilde X \in \R^{bk \times m}$ and augmented matricies $\tilde R \in \R^{bk \times m}$ and $\tilde S \in \R^{bk \times n}$ given by
\begin{align}
\tilde X = \begin{bmatrix} X \\ X \\ \vdots \\ X \end{bmatrix}, \ \ \ \ \ \tilde R =\begin{bmatrix} R^b_1 \\ R^b_2 \\ \vdots \\ R^b_k \end{bmatrix},  \ \ \ \ \   \tilde S =\begin{bmatrix} S^b_1 \\ S^b_2 \\ \vdots \\ S^b_k \end{bmatrix} .
\end{align}
Outputs are then computed as
\begin{align} \label{eq:bvabe}
    f(X, W) = \texttt{softmax}\round{ \round{\round{\tilde X \odot \tilde R }  W} \odot \tilde S }
\end{align}
where the $b$ rows $(bi,...,bi + b-1)$ of the output correspond exactly to Equation~\ref{eq:bvbe}. The task may then be inferred by choosing the $i$ for which the rows $(bi,...,b(i+1) - 1)$ minimize the objective $\Hcal$. If $f(X, W)_i$ denotes row $i$ of $f(X, W)$ then for objective $\Hcal$ the inferred task for ABatchE is
\begin{align}
    \argmin_i \sum_{\omega = 0}^{b-1} \Hcal\round{f(X, W)_{bi + \omega}}.
\end{align}
To extend ABatchE to deep neural networks the matricies $\tilde R$ and $\tilde S$ are constructed for each layer.

One advantage of ABatchE over \ac is that no backwards pass is required. However, ABatchE uses a very large batch size for large $k$, and the forward pass therefore requires more compute and memory. 
Another disadvantage of ABatchE is that the performance of ABatchE is limited by the performance of BatchE. In Section~\ref{sec:exp-s23} we demonstrate that \ac outperforms BatchE when BatchE is given task identity information.

\begin{figure}[t]
    \centering
    \includegraphics[width=\textwidth]{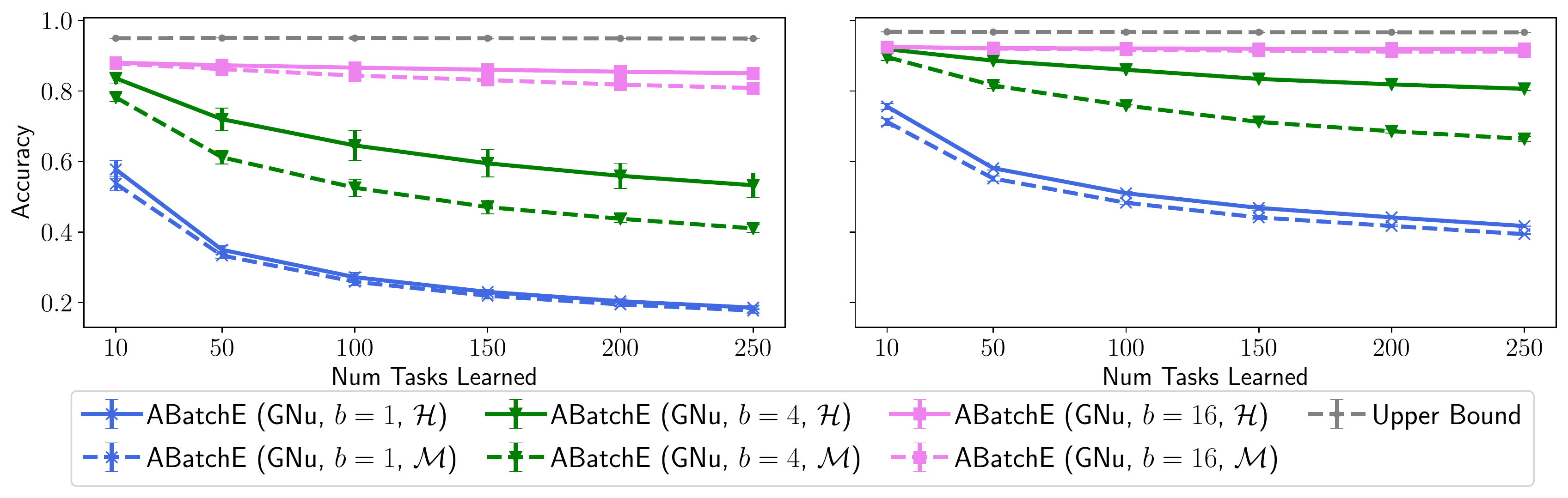}
    \caption{Testing ABatchE on PermutedMNIST with LeNet 300-100 \textbf{(left)} and FC 1024-1024 \textbf{(right)} with output size 100.}
    \label{fig:abe}
    \vspace{-1.5em}
\end{figure}

Since the objective for ABatchE need not be differentiable we also experiment with an alternative metric of confidence $\Mcal(\p) = -\max_i \p_i$.
We showcase results for ABatchE on PermutedMNIST in \autoref{fig:abe} for various values of $b$. The entropy objective $\Hcal$ performs better than $\Mcal$, and forgetting is only mitigated when using 16 images ($b=16$). With 250 tasks, $b=16$ corresponds to a batch size of 4000.

\section{Extended Training Details} \label{sec:hyperparams}
\subsection{SplitCIFAR-100 (\casename{GG})}\label{sec:gg}

As in \cite{wen2020batchensemble} we train each model for 250 epochs per task. We use standard hyperparameters---the Adam optimizer \cite{kingma2014adam} with a batch size of 128 and learning rate 0.001 (no warmup, cosine decay \cite{cosine}). For \ac we follow \cite{ramanujan2019s} and use non-affine normalization so there are no learned parameters. We do have to store the running mean and variance for each task, which we include in the parameter count. We found it better to use a higher learning rate (0.1) when training BatchE (Rand $W$), and the standard BatchE number is taken from \cite{wen2020batchensemble}.

\subsection{SplitImageNet (\casename{GG})}
We use the Upper Bound and BatchE number from \cite{wen2020batchensemble}. For \ac we train for 100 epochs with a batch size of 256 using the Adam optimizer \cite{kingma2014adam} with learning rate 0.001 (5 epochs warmup, cosine decay \cite{cosine}). For \ac we follow \cite{ramanujan2019s} and use non-affine normalization so there are no learned parameters. We do have to store the running mean and variance for each task, which we include in the parameter count.

\subsection{\casename{GNu} Experiments}
We clarify some experimental details for \casename{GNu} experiments \& baselines. For the BatchE \cite{wen2020batchensemble} baseline we find it best to use kaiming normal initialization with a learning rate of 0.01 (0.0001 for the first task when the weights are trained). As we are considering hundreds of tasks, instead of training separate heads per tasks when training BatchE we also apply the rank one pertubation to the final layer. PSP \cite{cheung2019superposition} provides MNISTPerm results so we use the same hyperparameters as in their code. We compare with rotational superposition, the best performing model from PSP.

\subsection{Speed of the Masked Forward Pass}\label{sec}

We now provide justification for the calculation mentioned in Section~\ref{sec:exp-s1}---when implemented properly the masking operation should require $\sim 1\%$ of the total time for a forward pass (for a ResNet-50 on a NVIDIA GTX 1080 Ti GPU). It is reasonable to assume that selecting indices is roughly as quick as memory access. A NVIDIA GTX 1080 Ti has a memory bandwidth of 480 GB/s. A ResNet-50 has around $2.5\cdot 10^7$ 4-byte (32-bit) parameters---roughly 0.1 GB. Therefore, indexing over a ResNet-50 requires at most $0.1 \text{ GB} / \left(480\text{ GB/s}\right) \approx 0.21\text{ ms}$. For comparison, the average forward pass of a ResNet-50 for a $3\times 224\times 224$ image on the same GPU is about 25 ms. 

Note that NVIDIA hardware specifications generally assume best-case performance with sequential page reads. However, even if real-world memory bandwidth speeds are 60-70\% slower than advertised, the fraction of masking time would remain in the $\leq 3$\% range.

\subsection{Additional Transfer Experiment}

For our transfer experiments, we initialize the score matrix (see \autoref{sec:supermask-training}) for task $i$ with the running mean of the supermasks for tasks 0 through $i-1$. The scores for task 0 are initialized as in~\cite{ramanujan2019s}. We further normalize by the Kaiming fan-in constant from \cite{he2015delving}, so that the norm of our supermask matrix is reasonable. If we do not perform this normalization, accuracy degrades significantly. All other training hyperparameters are the same as in Section~\ref{sec:gg}. 

\begin{figure}
    \begin{subfigure}{0.52\textwidth}
    \centering
    \includegraphics[scale=0.43]{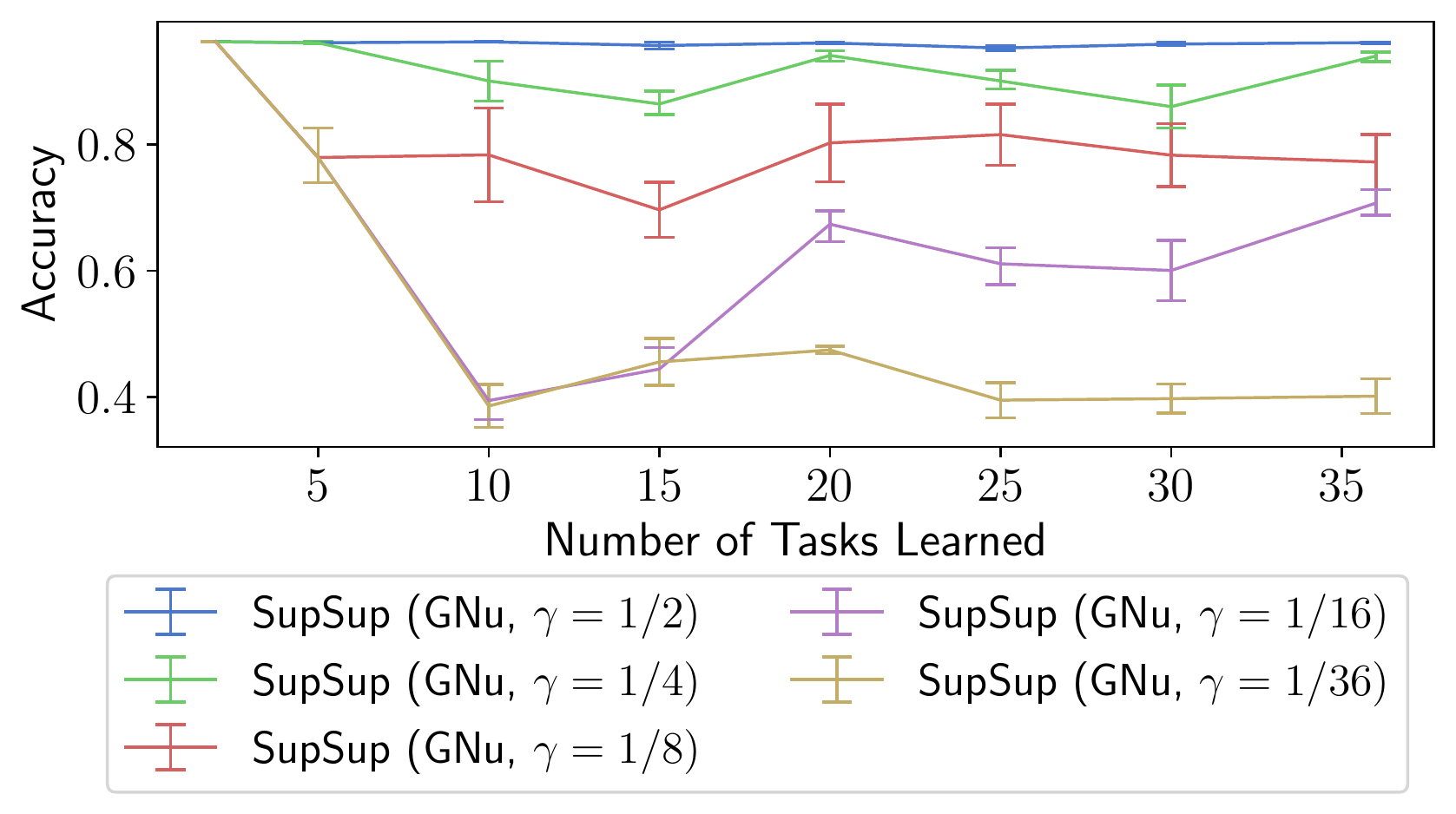}
    \end{subfigure}
    \begin{subfigure}{0.48\textwidth}
    \centering
    \includegraphics[scale=0.47]{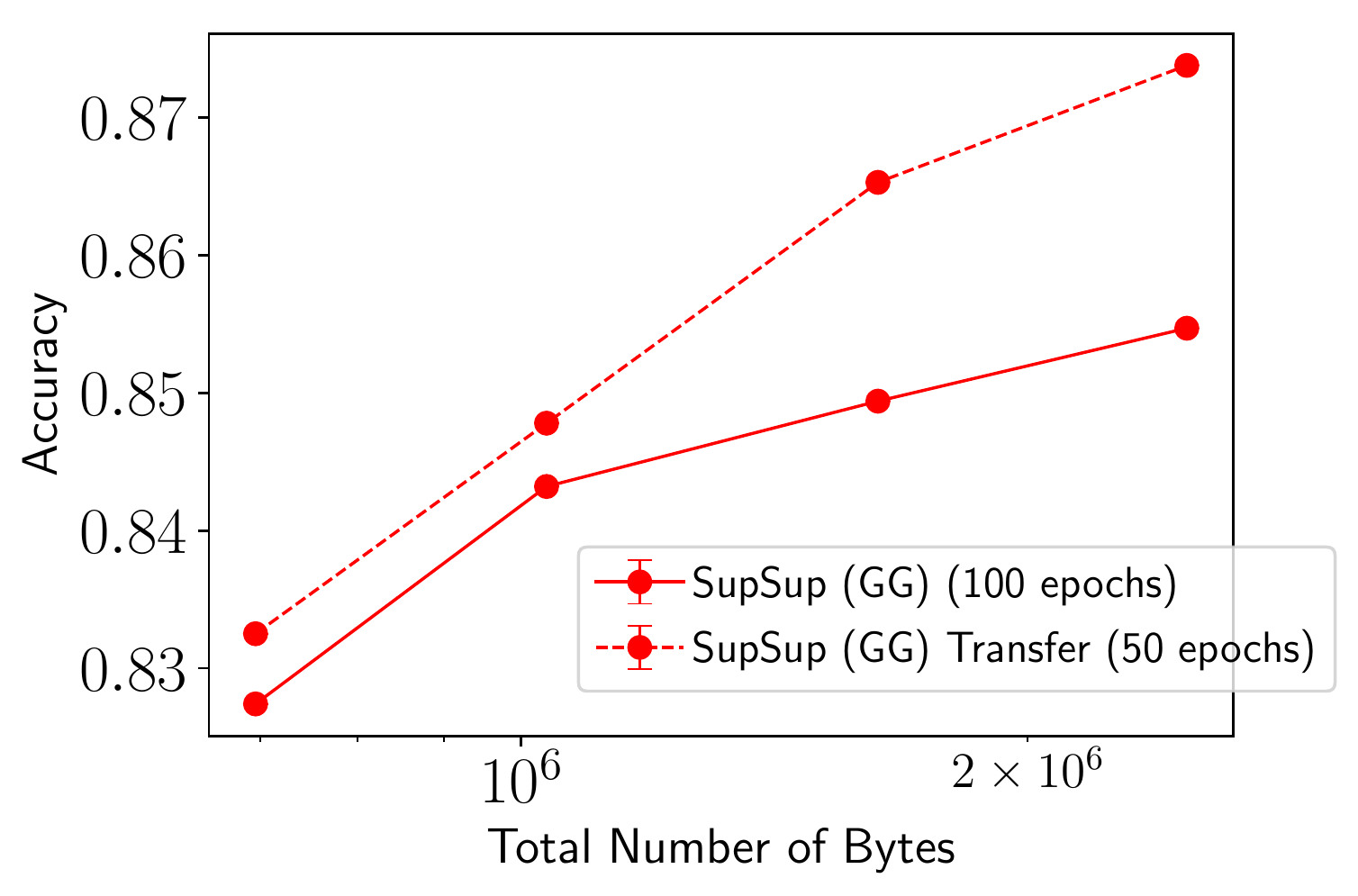}
    \end{subfigure}
    \caption{\textbf{(left)} Interpolating between the binary and one-shot algorithm with $\gamma$. \textbf{(right)} Transfer enables faster learning on SplitCIFAR.}
         \label{fig:rf}
    \vspace*{-1em}
\end{figure}

In \autoref{fig:rf}, we demonstrate that Transfer enables faster learning for SplitCIFAR. In this experiment, we train task 0 for the full 250 epochs and all subsequent tasks for either 50 epochs (with transfer) or 100 epochs (without transfer). We see that adding transfer yields an improvement even while using about half the number of training iterations overall.

\section{Supermask Training with Edge-Popup} \label{sec:supermask-training}
For completeness we briefly recap the Edge-Popup algorithm for training supermasks as introduced by \cite{ramanujan2019s}. Consider a linear layer with inputs $\textbf{x} \in \R^m$ and outputs $\textbf{y} = (W \odot M)^\top \textbf{x}$ where $W \in \R^{m \times n}$ are the fixed weights and $M \in \{0, 1\}^{m \times n}$ is the supermask. The Edge-Popup algorithm learns a score matrix $S \in \R_+^{m \times n}$ and computes the mask via $M = h(S)$. The function $h$ sets the top $k$\% of entries in $S$ to 1 and the remaining to 0. Edge-Popup updates $S$ via the straight through estimator---$h$ is considered to be the identity on the backwards pass.

\section{Comparing Binary and One-Shot}
In Figure~\ref{fig:rf} \textbf{(left)} we interpolate between the \textbf{Binary} and \textbf{One-Shot} algorithms. We replace line 6 of Algorithm~\ref{alg:binary}, $g_i \leq {\textbf{median}(g)}$, with $g_i \leq {\textbf{top-}\gamma\%\textbf{-element}(g)}$. Then when $\gamma=1/2$ we recover the binary algorithm (as $\textbf{median}(g) = \textbf{top-}50\%\textbf{-element}(g)$) and when $\gamma=1/k$ we recover the one-shot algorithm. A performance drop is observed from binary to one-shot for the difficult task of MNISTRotate---sequentially learning 36 rotations of MNIST (each rotation differing by 10 degrees).

\section{Tree Representation for the Continual Learning Scenarios}
In Figure~\ref{fig:scenarios} the Continual Learning scenarios are represented as a tree. This resembles the formulation from \cite{zeno2018task} with some modifications, \textit{i.e.} ``Tasks share output head?'' is replaced with ``Tasks share labels'' as it is possible to share the output head but not labels, \textit{e.g.} \ac in \casename{GNu}.

\section{Corresponding Tables}\label{sec:tables}
In this section we provide tabular results for figures from Section~\ref{sec:exps}.
\begin{table}[H]
    \centering
    \caption{Accuracy on SplitCIFAR100 corresponding to \autoref{fig:t1} \textbf{(right)}. \ac with Transfer approaches the upper bound.}
    \begin{tabular}{llr}
\toprule
                                          Entry &         Avg Acc@1 &    Bytes \\
\midrule
SupSup ($\ensuremath{\mathsf{GG}}$) &  $77.56~\pm 0.73$ &    408432 \\
SupSup ($\ensuremath{\mathsf{GG}}$) &  $83.62~\pm 0.74$ &    508432 \\
SupSup ($\ensuremath{\mathsf{GG}}$) &  $86.45~\pm 0.61$ &    695592 \\
SupSup ($\ensuremath{\mathsf{GG}}$) &  $88.09~\pm 0.64$ &   1035792 \\
SupSup ($\ensuremath{\mathsf{GG}}$) &  $89.06~\pm 0.75$ &   1630032 \\
SupSup ($\ensuremath{\mathsf{GG}}$) &  $89.57~\pm 0.64$ &   2487472 \\
SupSup ($\ensuremath{\mathsf{GG}}$) Transfer &  $79.53~\pm 1.31$ &    408432 \\
SupSup ($\ensuremath{\mathsf{GG}}$) Transfer &  $85.33~\pm 1.05$ &    508432 \\
SupSup ($\ensuremath{\mathsf{GG}}$) Transfer &  $88.52~\pm 0.85$ &    695592 \\
SupSup ($\ensuremath{\mathsf{GG}}$) Transfer &  $90.12~\pm 0.75$ &   1035792 \\
SupSup ($\ensuremath{\mathsf{GG}}$) Transfer &  $91.31~\pm 0.74$ &   1630032 \\
SupSup ($\ensuremath{\mathsf{GG}}$) Transfer &  \textbf{91.66} $~\pm 0.74$ &   2487472 \\
BatchE ($\ensuremath{\mathsf{GG}}$) &  $79.75~\pm 1.00$ &   4640800 \\
BatchE ($\ensuremath{\mathsf{GG}}$) - Rand $W$ &  $74.96~\pm 0.68$ &    400240 \\
Separate Heads &  $70.60~\pm 1.40$ &   4544560 \\
Separate Heads - Rand $W$ &  $50.00~\pm 1.37$ &    184000 \\\midrule
Upper Bound &  $91.62~\pm 0.89$ &  89675200 \\
\bottomrule
\end{tabular}

    \label{tab:t1}
\end{table}

\begin{table}[H]
    \centering
    \caption{Accuracy on  PermutedMNIST with LeNet 300-100 corresponding to \autoref{fig:v1-v2} \textbf{(left)}.}
    \begin{tabular}{lrrrrrrr}
\toprule
                                              Entry &    10 &    50 &   100 &   150 &   200 &   250 &   Avg \\
\midrule
 SupSup ($\ensuremath{\mathsf{GNu}}$ $\mathcal{H}$) & 93.65 & \textbf{93.68} & \textbf{93.68} & \textbf{93.66} & 93.64 & 93.62 & \textbf{93.66} \\
 SupSup ($\ensuremath{\mathsf{GNu}}$ $\mathcal{G}$) & 93.69 & 93.67 & 93.67 & \textbf{93.66} & \textbf{93.65} & \textbf{93.63} & \textbf{93.66} \\
                   PSP ($\ensuremath{\mathsf{GG}}$) & \textbf{94.80} & 83.58 & 64.62 & 51.18 & 42.69 & 36.74 & 62.27 \\
                BatchE ($\ensuremath{\mathsf{GG}}$) & 88.85 & 88.33 & 88.23 & 88.23 & 88.22 & 88.21 & 88.34 \\
                \midrule
                                        Upper Bound & 94.94 & 95.01 & 94.99 & 94.95 & 94.91 & 94.86 & 94.94 \\
\bottomrule
\end{tabular}

    \label{tab:v1-v2-left}
\end{table}
\begin{table}[H]
    \centering
        \caption{Accuracy on PermutedMNIST with FC 1024-1024 corresponding to \autoref{fig:v1-v2} \textbf{(right)}.}
    \begin{tabular}{lrrrrrrr}
\toprule
                                              Entry &    10 &    50 &   100 &   150 &   200 &   250 &   Avg \\
\midrule
 SupSup ($\ensuremath{\mathsf{GNu}}$ $\mathcal{H}$) & 96.28 & 96.14 & 96.04 & 95.91 & 95.86 & 95.66 & 95.98 \\
 SupSup ($\ensuremath{\mathsf{GNu}}$ $\mathcal{G}$) & 96.28 & \textbf{96.19} & \textbf{96.05} & \textbf{96.00} & \textbf{95.99} & \textbf{95.92} & \textbf{96.07} \\
                   PSP ($\ensuremath{\mathsf{GG}}$) & \textbf{97.16} & 94.74 & 87.77 & 78.35 & 69.14 & 61.11 & 81.38 \\
                BatchE ($\ensuremath{\mathsf{GG}}$) & 92.84 & 92.40 & 92.36 & 92.34 & 92.33 & 92.32 & 92.43 \\
                \midrule
                                        Upper Bound & 96.76 & 96.70 & 96.68 & 96.66 & 96.63 & 96.61 & 96.67 \\
\bottomrule
\end{tabular}

    \label{tab:v1-v2-right}
\end{table}
\begin{table}[H]
    \centering
    \caption{Accuracy on PermutedMNIST with LeNet 300-100 corresponding to \autoref{fig:long-zoom}.}
    \begin{tabular}{lrrrrrr}
\toprule
                                              Entry &   500 &  1000 &  1500 &  2000 &  2500 &   Avg \\
\midrule
 SupSup ($\ensuremath{\mathsf{GNu}}$ $\mathcal{H}$) & 93.49 & 93.47 & 93.46 & 93.45 & 93.45 & 93.46 \\
 SupSup ($\ensuremath{\mathsf{GNu}}$ $\mathcal{G}$) & 93.49 & 93.48 & 93.46 & 93.45 & 93.45 & 93.47 \\
 SupSup ($\ensuremath{\mathsf{NNs}}$ $\mathcal{H}$) & 93.49 & 93.46 & 93.46 & 93.45 & 92.54 & 93.28 \\\midrule
                                        Upper Bound & 94.71 & 94.71 & 94.71 & 94.71 & 94.71 & 94.71 \\
\bottomrule
\end{tabular}

    \label{tab:long}
\end{table}
\begin{table}[H]
    \centering
    \caption{Accuracy with FC 1024-1024 on RotatedMNIST corresponding to \autoref{fig:rot-adapt} \textbf{(left)}.}
    \begin{tabular}{lr}
\toprule
                                                         Entry &   Avg \\
\midrule
 SupSup ($\ensuremath{\mathsf{GNu}}$ full batch $\mathcal{H}$) & \textbf{96.13} \\
                           BatchE ($\ensuremath{\mathsf{GG}}$) & 92.40 \\
                              PSP ($\ensuremath{\mathsf{GG}}$) & 95.87 \\
                                                   Lower Bound & 48.71 \\\midrule
                                                   Upper Bound & 98.01 \\
\bottomrule
\end{tabular}

    \label{tab:rot-adapt-left}
\end{table}
\begin{table}[H]
    \centering
    \caption{Accuracy with FC 1024-1024 on PermutedMNIST corresponding to \autoref{fig:rot-adapt} \textbf{(right)}.}
    \begin{tabular}{lrrrrrrr}
\toprule
                                                         Entry &    10 &    50 &   100 &   150 &   200 &   250 &   Avg \\
\midrule
            SupSup ($\ensuremath{\mathsf{GNu}}$ $\mathcal{H}$) & \textbf{96.29} & \textbf{95.94} & \textbf{95.59} & \textbf{95.40} & \textbf{95.00} & \textbf{94.91} & \textbf{95.52} \\
 BatchE ($\ensuremath{\mathsf{GNu}}$ full batch $\mathcal{H}$) & 91.94 & 91.90 & 92.04 & 92.04 & 92.04 & 92.04 & 92.00 \\
            BatchE ($\ensuremath{\mathsf{GNu}}$ $\mathcal{H}$) & 66.08 & 61.89 & 60.93 & 59.33 & 57.37 & 55.74 & 60.22 \\\midrule
                                                   Upper Bound & 96.76 & 96.70 & 96.68 & 96.66 & 96.63 & 96.61 & 96.67 \\
\bottomrule
\end{tabular}

    \label{tab:rot-adapt-right}
\end{table}
\begin{table}[H]
    \centering
    \caption{Accuracy on PermutedMNIST with LeNet 300-100 corresponding to \autoref{fig:v1-v2-sz} \textbf{(left)}.}
    \begin{tabular}{lrrrrrrr}
\toprule
                                                      Entry &    10 &    50 &   100 &   150 &   200 &   250 &   Avg \\
\midrule
 SupSup ($\ensuremath{\mathsf{GNu}}$ $s=200$ $\mathcal{H}$) & 93.46 & 93.49 & 93.48 & 93.47 & 93.47 & 93.46 & 93.47 \\
 SupSup ($\ensuremath{\mathsf{GNu}}$ $s=200$ $\mathcal{G}$) & 93.46 & 93.48 & 93.47 & 93.47 & 93.47 & 93.46 & 93.47 \\
 SupSup ($\ensuremath{\mathsf{GNu}}$ $s=100$ $\mathcal{H}$) & 93.65 & \textbf{93.68} & \textbf{93.68} & \textbf{93.66} & 93.64 & 93.62 & \textbf{93.66} \\
 SupSup ($\ensuremath{\mathsf{GNu}}$ $s=100$ $\mathcal{G}$) & \textbf{93.69} & 93.67 & 93.67 & \textbf{93.66} & \textbf{93.65} & \textbf{93.63} & \textbf{93.66} \\
  SupSup ($\ensuremath{\mathsf{GNu}}$ $s=25$ $\mathcal{H}$) & 93.71 & 93.51 & 93.28 & 93.10 & 93.06 & 92.94 & 93.27 \\
  SupSup ($\ensuremath{\mathsf{GNu}}$ $s=25$ $\mathcal{G}$) & 93.83 & 93.66 & 93.60 & 93.48 & 93.43 & 93.36 & 93.56 \\
                                                Lower Bound & 71.67 & 41.82 & 30.52 & 26.40 & 23.31 & 20.88 & 35.77 \\\midrule
                                                Upper Bound & 94.94 & 95.01 & 94.99 & 94.95 & 94.91 & 94.86 & 94.94 \\
\bottomrule
\end{tabular}

    \label{tab:v1-v2-sz-left}
\end{table}
\begin{table}[H]
    \centering
    \caption{Accuracy on PermutedMNIST with FC 1024-1024 corresponding to \autoref{fig:v1-v2-sz} \textbf{(right)}.}
    \begin{tabular}{lrrrrrrr}
\toprule
                                                      Entry &    10 &    50 &   100 &   150 &   200 &   250 &   Avg \\
\midrule
 SupSup ($\ensuremath{\mathsf{GNu}}$ $s=200$ $\mathcal{H}$) & 96.28 & 96.14 & 96.04 & 95.91 & 95.86 & 95.66 & 95.98 \\
 SupSup ($\ensuremath{\mathsf{GNu}}$ $s=200$ $\mathcal{G}$) & 96.28 & \textbf{96.19} & \textbf{96.05} & \textbf{96.00} & \textbf{95.99} & \textbf{95.92} & \textbf{96.07} \\
 SupSup ($\ensuremath{\mathsf{GNu}}$ $s=100$ $\mathcal{H}$) & 95.90 & 94.77 & 94.02 & 93.71 & 93.00 & 92.84 & 94.04 \\
 SupSup ($\ensuremath{\mathsf{GNu}}$ $s=100$ $\mathcal{G}$) & \textbf{96.31} & 95.83 & 95.60 & 95.32 & 95.05 & 94.88 & 95.50 \\
  SupSup ($\ensuremath{\mathsf{GNu}}$ $s=25$ $\mathcal{H}$) & 82.28 & 69.06 & 64.51 & 60.99 & 58.15 & 57.03 & 65.34 \\
  SupSup ($\ensuremath{\mathsf{GNu}}$ $s=25$ $\mathcal{G}$) & \textbf{96.31} & 93.17 & 91.20 & 90.26 & 89.04 & 88.19 & 91.36 \\
                                                Lower Bound & 76.89 & 49.40 & 38.93 & 34.53 & 31.30 & 29.36 & 43.40 \\\midrule
                                                Upper Bound & 96.76 & 96.70 & 96.68 & 96.66 & 96.63 & 96.61 & 96.67 \\
\bottomrule
\end{tabular}

    \label{tab:v1-v2-sz-right}
\end{table}

\section{Analysis} \label{sec:analysis}

In this section we assume a slightly more technical perspective. The aim is not to formally prove properties of the algorithm. Rather, we hope that a more mathematical language may prove useful in extending intuition. Just as the empirical work of \cite{frankle2018lottery, zhou2019deconstructing, ramanujan2019s} was given a formal treatment in \cite{malach2020proving}, we hope for more theoretical work to follow.

Our grounding intuition remains from Section~\ref{sec:S23}---the correct mask will produce the lowest entropy output. Moreover, since entropy is differentiable, gradient based optimization can be used to recover the correct mask. However, many questions remain: Why do superfluous neurons (Section~\ref{sec:s-neuron}) help? In the case of MNISTPermuation, why is a single gradient sufficient? Although it is a simple case, steps forward can be made by analyzing the training of a linear head on fixed features. With \textit{random} features, training a linear head on fixed features is considered in the literature of reservoir computing \cite{schrauwen2007overview}, and more \cite{bengio2006convex}.

Consider $k$ different classification problems with fixed features $\phi(\x) \in \R^m$. Traditionally, one would use learned weights $W \in \R^{m \times n}$ to compute \textit{logits}
\begin{align}
    \y = W^\top \phi(\x)
\end{align}
and output classification probabilities $\p = \texttt{softmax}(\y)$ where
\begin{align}
    \p_v = \frac{\exp(\y_v)}{\sum_{v' = 1}^n \exp(\y_{v'})}.
\end{align}
Recall that with \ac we compute the \textit{logits} for task $i$ using fixed random weights $W$ and a learned binary mask $M^i \in \{0,1\}^{m \times n}$ as
\begin{align}
    \y = \round{W \odot M^i}^\top \phi(\x)
\end{align}
where $\odot$ denotes an element-wise product and no bias term is allowed. Moreover, $W_{uv} =  \xi_{uv} \sqrt{{2} /{m}} $ where $\xi_{uv}$ is chosen independently to be either $-1$ or $1$ with equal probability and the constant $\sqrt{{2} /{m}}$ follows Kaiming initialization~\cite{he2015delving}.

Say we are given data $\x$ from task $j$. From now on we will refer to task $j$ as the \textit{correct} task. Recall from~\secref{S23} that \ac attempts to infer the \textit{correct} task by using a weighted mixture of masks
\begin{align} \label{eq:susu-elm}
    \y = \round{ W \odot \sum_i \alpha_i M^i}^\top \phi(\x)
\end{align}
where the coefficients $\alpha_i$ sum to one, and are initially set to $1/k$.

To infer the correct task we attempt to construct a function $\Gcal(\y; \alpha)$ with the following property: For fixed data, $\Gcal$ is minimized when $\alpha = \e_j$ ($\e_j$ denotes a $k$-length vector that is 1 in index $j$ and 0 otherwise). We can then infer the correct task by solving a minimization problem.

As in \textbf{One-Shot}, we use a single gradient computation to infer the task via
\begin{align}
\argmax_i  \round{ -\frac{\partial \Gcal}{\partial \alpha_i} }.
\end{align}
A series of Lemmas will reveal how a single gradient step may be sufficient when tasks are unrelated (\textit{e.g.} as in PermutedMNIST). We begin with the construction of a useful function $\Gcal$, which will correspond exactly to $\Gcal$ in Section~\ref{sec:s-neuron}. As in Section~\ref{sec:s-neuron}, this construction is made possible through superfluous neurons (s-neurons): The true labels are in $\{1,...,\ell \}$, and a typical output is therefore length $\ell$. However, we add $n-\ell$ s-neurons resulting in a vector $\y$ of length $n$. 

Let $\mathbf{S}$ denote the set of s-neurons and $\mathbf{R}$ denote the set of \emph{real} neurons where $|\mathbf{S}| = n-\ell$ and $|\mathbf{R}| = \ell$. Moreover, assume that a standard cross-entropy loss is used during training, which will encourage s-neurons to have small values. 

\begin{lemma} \label{lemma:constructg}
It is possible to construct a function $\Gcal$ such that the gradient matches the gradient from the supervised training loss $\Lcal$ for all s-neurons.  Specifically, $\frac{\partial \Gcal}{\partial y_v} =  \frac{\partial \Lcal}{\partial y_v}$ for all $v \in \mathbf{S}$ and 0 otherwise. 
\begin{proof}
Let $g_v = \frac{\partial \Gcal}{\partial y_v}$. It is easy to ensure that $g_v = 0$ for all $v \not\in \mathbf{S}$ with a modern neural network library like PyTorch \cite{paszke2019pytorch} as \emph{detaching}\footnote{\url{https://pytorch.org/docs/stable/autograd.html}} the outputs from the neurons $v \not\in \mathbf{S}$ prevents gradient signal from reaching them. In code, let \texttt{y} be the outputs and \texttt{m} be a binary vector with $\texttt{m}_v = 1$ if $v \in \mathbf{S}$ and 0 otherwise, then
\begin{align}
    \texttt{y = (1 - m) * y.detach() + m * y}
\end{align}
will prevent gradient signal from reaching $\y_v$ for $v \not\in \mathbf{S}$.

Recall that the standard cross-entropy loss is
\begin{align}
    \Lcal(\y) = -\log \round{\frac{\exp(\y_c)}{\sum_{v' = 1}^n \exp(\y_{v'})}}
    = - \y_c + \log\round{\sum_{v' = 1}^n \exp(\y_{v'})}
\end{align}
where $c \in \{1,...,\ell\}$ is the correct label. The gradient of $\Lcal$ to any s-neuron $v$ is then
\begin{align}
    \frac{\partial \Lcal}{\partial \y_v} = \frac{\exp(\y_v)}{\sum_{v' = 1}^n \exp(\y_{v'})}.
\end{align}
If we define $\Gcal$ as 
\begin{align}
    \Gcal(\y; \alpha) = \log\round{\sum_{v' = 1}^n \exp(\y_{v'})}
\end{align}
then $g_v =  \frac{\partial \Lcal}{\partial y_v}$ as needed. Expressed in code
\begin{align}
    &\texttt{y = model(x)}; \ \ \ \texttt{G = torch.logsumexp((1 - m) * y.detach() + m * y, dim=1)}
\end{align}
where \texttt{model(...)} computes Equation \ref{eq:susu-elm}.
\end{proof}
\end{lemma}
In the next two Lemmas we aim to show that, in expectation, $-\frac{\partial \Gcal}{\partial \alpha_i} \leq 0$ for $i \neq j$ while $-\frac{\partial \Gcal}{\partial \alpha_j} > 0$. Recall that $j$ is the \textit{correct} task---the task from which the data is drawn---and we will use $i$ to refer to a different task.

When we take expectation, it is with respect to the random variables $\xi, \{M^\omega\}_{\omega \in \{1,..,k\}}$, and $\x$. Before we proceed further a few assumptions are formalized, \textit{e.g.} what it means for tasks to be unrelated.

\textbf{Assumption 1:} We assume that the mask learned on task $i$ will be independent from the data from task $j$: If the data is from task $j$ then $\phi(\x)$ and $M^i$ and independent random variables.

\textbf{Assumption 2:} We assume that a negative weight and positive weight are equally likely to be masked out. As a result, $\E{\xi_{uv}M_{uv}^i} = 0$. Note that when $\E{\phi(\x)} = 0$, which will be the case for zero mean random features, there should be little doubt that this assumption should hold.

\begin{lemma}
If data $\x$ comes from task $j$ and $i \neq j$ then
\begin{align}
    \E{-\frac{\partial \Gcal}{\partial \alpha_i} } \leq 0
\end{align}
\begin{proof}
We may write the gradient as
\begin{align} \label{eq:lemgrad}
    \frac{\partial \Gcal}{\partial \alpha_i} = \sum_{v=1}^n \frac{\partial \Gcal}{\partial \y_v} \frac{\partial \y_v}{\partial \alpha_i} 
\end{align}
and use that $\frac{\partial \Gcal}{\partial \y_v} = 0$ for $v \not \in \mathbf{S}$. Moreover, $\y_v$ may be written as
\begin{align}
    \y_v = \sum_{u = 1}^n \phi(\x)_u W_{uv} \round{\sum_{i=1}^k \alpha_i M^i_{uv}}
\end{align}
with $W_{uv} = \xi_{uv} \sqrt{2 / m}$ and so Equation \ref{eq:lemgrad} becomes
\begin{align}
    \frac{\partial \Gcal}{\partial \alpha_i} = \frac{\sqrt 2}{\sqrt m} \sum_{v \in \mathbf{S}}  \sum_{u = 1}^n   \frac{\partial \Gcal}{\partial \y_v} \phi(\x)_u   \xi_{uv} M_{uv}^i.
\end{align}
Taking the expectation (and using linearity) we obtain
\begin{align} \label{eq:grad-simple}
    \E{\frac{\partial \Gcal}{\partial \alpha_i}} = \frac{\sqrt 2}{\sqrt m} \sum_{v \in \mathbf{S}}  \sum_{u = 1}^n   \E{ \frac{\partial \Gcal}{\partial \y_v} \phi(\x)_u   \xi_{uv} M_{uv}^i }.
\end{align}
In Lemma \ref{lem:tech} we formally show that each term in this sum is greater than or equal to 0, which completes this proof. However, we can see informally now why expectation should be close to 0 if we ignore the gradient term as
\begin{align}
   \E{ \phi(\x)_u   \xi_{uv} M_{uv}^i } = \E{ \phi(\x)_u  } \E{ \xi_{uv} M_{uv}^i } = 0
\end{align}
where the first equality follows from Assumption 1 and the latter follows from Assumption 2.
\end{proof}
\end{lemma}

We have now seen that in expectation $-\frac{\partial \Gcal}{\partial \alpha_i} \leq 0$ for $i \neq j$. It remains to be shown that we should expect $-\frac{\partial \Gcal}{\partial \alpha_j} > 0$.

\begin{lemma}
If data $\x$ comes from the task $j$ then
\begin{align}
    \E{-\frac{\partial \Gcal}{\partial \alpha_j}} > 0.
\end{align}
\begin{proof}
Following Equation \ref{eq:grad-simple}, it suffices to show that for $u \in \{1,...,m\}$, $v \in \mathbf{S}$
\begin{align} 
    \E{ -\frac{\partial \Gcal}{\partial \y_v} \phi(\x)_u   \xi_{uv} M_{uv}^j } > 0.
\end{align}
Since $v \in \mathbf{S}$ we may invoke Lemma \ref{lemma:constructg} to rewrite our objective as
\begin{align} \label{eq:lem-new-obj}
    \E{- \frac{\partial \Lcal}{\partial \y_v} \phi(\x)_u   \xi_{uv} M_{uv}^j } > 0
\end{align}
where $\Lcal$ is the supervised loss used for training. Recall that in the mask training algorithm, real valued scores $S_{uv}^j$ are associated with $M_{uv}^j$ \cite{ramanujan2019s, mallya2018piggyback}. The update rule for $S_{uv}^j$ on the backward pass is then
\begin{align}
    S_{uv}^j \gets S_{uv}^j + \eta \round{- \frac{\partial \Lcal}{\partial \y_v} \phi(\x)_u   \xi_{uv} }
\end{align}
for some learning rate $\eta$. Following Mallya \textit{et al.} \cite{mallya2018piggyback} (with threshold 0, as used in Section~\ref{sec:exp-s23}), we let $M_{uv}^j=1$ if $S_{uv}^j > 0$ and otherwise assign $M_{uv}^j=0$. As a result, we expect that $M_{uv}^j$ is 1 when $- \frac{\partial \Lcal}{\partial \y_v} \phi(\x)_u   \xi_{uv} $ is more consistently positive than negative. In other words, the expected product of $M_{uv}^j$ and $- \frac{\partial \Lcal}{\partial \y_v} \phi(\x)_u   \xi_{uv} $ is positive, satisfying Equation \ref{eq:lem-new-obj}.
\end{proof}
\end{lemma}
Together, three Lemmas have demonstrated that in expectation $-\frac{\partial \Gcal}{\partial \alpha_i} \leq 0$ for $i \neq j$ while $-\frac{\partial \Gcal}{\partial \alpha_j} > 0$. Accordingly, we should expect that
\begin{align}
\argmax_i  \round{ -\frac{\partial \Gcal}{\partial \alpha_i} }.
\end{align}
returns the correct task $j$. While a full, formal treatment which includes the analysis of noise is beyond the scope of this work, we hope that this section has helped to further intuition. However, we are missing one final piece---what is the relation between $\Gcal$ and $\Hcal$?

It is not difficult to imagine that $\Hcal$ should imitate the loss, which attempts to raise the score of one logit while bringing all others down. Analytically we find that $\Hcal$ can be decomposed into two terms as follows
\begin{align}
    \Hcal\round{\p} &= -\sum_{v=1}^n \p_v \log \p_v \\
    &= -\sum_{v=1}^n \p_v \log \round{\frac{\exp\round{\y_v}}{\sum_{v'=1}^n \exp\round{\y_v'} }} \\
    &= \round{-\sum_{v=1}^n \p_v \y_v} + \log \round{\sum_{v'=1}^n \exp\round{\y_v'} }
\end{align}
where the latter term is $\Gcal$. With more and more neurons in the output layer, $\p_v$ will become small moving $\Hcal$ towards $\Gcal$.
\section{Additional Technical Details}

\begin{lemma} \label{lem:tech}
If $j$ is the true task and $i \neq j$ then
\begin{align} \label{eq:technical}
    \E{\frac{\partial \Gcal}{\partial \y_v} \phi(\x)_u   \xi_{uv} M_{uv}^i} \geq 0
\end{align}
\begin{proof}
Recall from Lemma \ref{lemma:constructg} that 
\begin{align}
    \frac{\partial \Gcal}{\partial \y_v} = \p_v = \frac{\exp(\y_v)}{\sum_{v' = 1}^n \exp(\y_{v'}) }
\end{align}
and so we rewrite equation \ref{eq:technical} as
\begin{align}
    \E{\p_v \phi(\x)_u   \xi_{uv} M_{uv}^i} \geq 0.
\end{align}
By the law of total expectation
\begin{align}
   \E{\p_v \phi(\x)_u   \xi_{uv} M_{uv}^i} =  \E{ \E{\p_v \phi(\x)_u   \xi_{uv} M_{uv}^i \Bigg| \left| \phi(\x)_u   \xi_{uv} M_{uv}^i \right| } }
\end{align}
and so it suffices to show that
\begin{align} \label{eq:suffices1}
    \E{\p_v \phi(\x)_u   \xi_{uv} M_{uv}^i \Bigg| \left| \phi(\x)_u   \xi_{uv} M_{uv}^i \right| = \kappa } \geq 0
\end{align}
for any $\kappa \geq 0$. In the case where where $\kappa = 0$ Equation \ref{eq:suffices1} becomes
\begin{align} \label{eq:suffices2}
    \E{ 0 \p_v \Bigg| \left| \phi(\x)_u   \xi_{uv} M_{uv}^i \right| = 0 } = 0
\end{align}
and so we are only left to consider $\kappa > 0$. Note that $\kappa > 0$ restricts $M_{uv}^i$ to be 1.

Again invoking the law of total expectation we rewrite Equation \ref{eq:suffices2} as
\begin{align}
\begin{split}
    &\E{\p_v \phi(\x)_u   \xi_{uv} M_{uv}^i \Bigg| \left| \phi(\x)_u   \xi_{uv} M_{uv}^i \right| } \\
    &= \E{\p_v \phi(\x)_u   \xi_{uv} M_{uv}^i \Bigg|  \phi(\x)_u   \xi_{uv} M_{uv}^i = \kappa }\PR{\phi(\x)_u   \xi_{uv} M_{uv}^i = \kappa} \\
    &+ \E{\p_v \phi(\x)_u   \xi_{uv} M_{uv}^i \Bigg|  \phi(\x)_u   \xi_{uv} M_{uv}^i = -\kappa }\PR{\phi(\x)_u   \xi_{uv} M_{uv}^i = -\kappa}.
    \end{split} \label{eq:lawtotalexp2}
\end{align}
Moreover, since the data is from task $j \neq i$, we can use Assumption 1 and 2 to show that each of the cases above is equally likely. Formally,
\begin{align}
    &\PR{\phi(\x)_u \xi_{uv} M_{uv}^i = \kappa}  \\
    &= \PR{ \round{\curly{\phi(\x)_u = \kappa} \cap \curly{ \xi_{uv} M_{uv}^i = 1 }} \cup \round{\curly{ \phi(\x)_u  = -\kappa } \cap \curly{ \xi_{uv} M_{uv}^i = -1 }}} \\
    &= \PR{\phi(\x)_u = \kappa } \PR{  \xi_{uv} M_{uv}^i = +1 } + \PR{ \phi(\x)_u  = -\kappa } \PR{ \xi_{uv} M_{uv}^i = -1 } \\
    &= \PR{\phi(\x)_u = \kappa } \PR{  \xi_{uv} M_{uv}^i = -1 } + \PR{ \phi(\x)_u  = -\kappa } \PR{ \xi_{uv} M_{uv}^i = +1 } \\
    &= \PR{ \round{\curly{\phi(\x)_u = \kappa} \cap \curly{ \xi_{uv} M_{uv}^i = -1 }} \cup \round{\curly{ \phi(\x)_u  = -\kappa } \cap \curly{ \xi_{uv} M_{uv}^i = +1 }}} \\
    &=\PR{\phi(\x)_u \xi_{uv} M_{uv}^i = -\kappa} 
\end{align}
and so we may factor out the probability terms in Equation \ref{eq:lawtotalexp2}. Accordingly, it suffices to show that
\begin{align} \label{eq:final-obj}
    \E{\p_v \phi(\x)_u   \xi_{uv} M_{uv}^i \Bigg|  \phi(\x)_u   \xi_{uv} M_{uv}^i = \kappa } + \E{\p_v \phi(\x)_u   \xi_{uv} M_{uv}^i \Bigg|  \phi(\x)_u   \xi_{uv} M_{uv}^i = -\kappa } \geq 0.
\end{align}
Before we proceed, we will introduce a function $h$ which we use to denote
\begin{align}
    h\round{\{\y_v\}, \kappa} = \kappa \frac{\exp(\y_v + \kappa)}{\exp(\y_v + \kappa) + \sum_{v' \neq v}\exp(\y_{v'})}.
\end{align}
for $\kappa > 0$. We will make use of two interesting properties of $h$.

We first note that $h\round{\{\y_v\}, \kappa} + h\round{\{\y_v\}, -\kappa} \geq 0$, which is formally shown in \ref{lemma:hprop1}.

Second, we note that
\begin{align}
\begin{split}
    &\PR{\p_v \phi(\x)_u   \xi_{uv} M_{uv}^i  = h\round{\{\y_v\}, \kappa} | \phi(\x)_u   \xi_{uv} M_{uv}^i = \kappa } \\ &= \PR{\p_v \phi(\x)_u   \xi_{uv} M_{uv}^i  = h\round{\{\y_v\}, -\kappa} | \phi(\x)_u   \xi_{uv} M_{uv}^i = -\kappa}
    \end{split}
\end{align}
which we dissect in Lemma \ref{lemma:hprop2}.

Utilizing these two properties of $h$ we may show that Equation \ref{eq:final-obj} holds as
\begin{align}
    &\E{\p_v \phi(\x)_u   \xi_{uv} M_{uv}^i \Bigg|  \phi(\x)_u   \xi_{uv} M_{uv}^i = \kappa } + \E{\p_v \phi(\x)_u   \xi_{uv} M_{uv}^i \Bigg|  \phi(\x)_u   \xi_{uv} M_{uv}^i = -\kappa } \\
    \begin{split}
    &= \int_\R h\round{\{\y_v\}, \kappa}  \ d\PR{ \p_v \phi(\x)_u   \xi_{uv} M_{uv}^i = h\round{\{\y_v\}, \kappa} | \phi(\x)_u   \xi_{uv} M_{uv}^i = \kappa } \\
    &\ \ + \int_\R h\round{\{\y_v\}, -\kappa}  \ d\PR{ \p_v \phi(\x)_u   \xi_{uv} M_{uv}^i = h\round{\{\y_v\}, -\kappa} | \phi(\x)_u   \xi_{uv} M_{uv}^i = -\kappa } 
    \end{split}\\
    &= \int_\R  \round{ h\round{\{\y_v\}, \kappa} + h\round{\{\y_v\}, -\kappa} }  \ d\PR{ \p_v \phi(\x)_u   \xi_{uv} M_{uv}^i = h\round{\{\y_v\}, \kappa} | \phi(\x)_u   \xi_{uv} M_{uv}^i = \kappa } \\
    &\geq 0.
\end{align}
\end{proof}
\end{lemma}

\begin{lemma}\label{lemma:hprop1} $h\round{\{\y_v\}, \kappa} + h\round{\{\y_v\}, -\kappa} \geq 0$.

\begin{proof}
Recall that $\kappa \geq 0$. Moreover,
\begin{align}
    &\exp(\y_v + \kappa) \sum_{v'} \exp(\y_{v'}) \geq \exp(\y_v - \kappa)\sum_{v'}  \exp(\y_{v'}) \\
    \begin{split}
    &\Rightarrow \ \exp(\y_v + \kappa) \round{\exp(\y_v - \kappa) + \sum_{v'} \exp(\y_{v'})}
    \\ &\ \ \ \ \ \ \geq \exp(\y_v - \kappa)\round{ \exp(\y_v + \kappa) + \sum_{v'} \exp(\y_{v'})}
    \end{split} \\
    &\Rightarrow \kappa \frac{\exp(\y_v + \kappa)}{\exp(\y_v + \kappa) + \sum_{v' \neq v}\exp(\y_{v'})} 
    \geq \kappa \frac{\exp(\y_v - \kappa)}{\exp(\y_v - \kappa) + \sum_{v' \neq v}\exp(\y_{v'})}
\end{align}
and we may then subtract the term on the right from both sides.
\end{proof}
\end{lemma}

\begin{lemma}\label{lemma:hprop2}
Consider take $i \neq j$ where $j$ is the correct task. Then
\begin{align}
    \begin{split} \label{eq:lemmab3}
        &\PR{\p_v \phi(\x)_u   \xi_{uv} M_{uv}^i  = h\round{\{\y_v\}, \kappa} | \phi(\x)_u   \xi_{uv} M_{uv}^i = \kappa } \\ &= \PR{\p_v \phi(\x)_u   \xi_{uv} M_{uv}^i  = h\round{\{\y_v\}, -\kappa} | \phi(\x)_u   \xi_{uv} M_{uv}^i = -\kappa}.
    \end{split}
\end{align}
\begin{proof}
Note that this equation is satisfied when $\kappa = 0$ (since $-0 = 0$). For the remainder of this proof we will instead consider the case where $\kappa > 0$ (and so $M_{uv}^i = 1$). 

If we define $\rho$ as $\rho = \round{\PR{\phi(\x)_u = \kappa} + \PR{\phi(\x)_u = -\kappa}}^{-1}$
then may decompose Equation \ref{eq:lemmab3} into four terms. Namely,
\begin{align}
\begin{split}
    & \PR{\p_v \phi(\x)_u   \xi_{uv} M_{uv}^i  = h\round{\{\y_v\},  \kappa} | \phi(\x)_u = \kappa } \PR{\phi(\x)_u = \kappa}\rho \\ 
    &+ \PR{\p_v \phi(\x)_u   \xi_{uv} M_{uv}^i  = h\round{\{\y_v\},  \kappa} | \phi(\x)_u = -\kappa } \PR{\phi(\x)_u = -\kappa} \rho\\
    &=\PR{\p_v \phi(\x)_u   \xi_{uv} M_{uv}^i  = h\round{\{\y_v\},  -\kappa} | \phi(\x)_u = \kappa } \PR{\phi(\x)_u = \kappa}\rho \\ 
    &+ \PR{\p_v \phi(\x)_u   \xi_{uv} M_{uv}^i  = h\round{\{\y_v\},  -\kappa} | \phi(\x)_u = -\kappa } \PR{\phi(\x)_u = -\kappa} \rho.
\end{split}
\end{align}
Equality follows from the fact that term 1 and 3 are equal, as are terms 2 and 4. We will consider terms 1 and 3, as the other case is nearly identical.

Let $H$ be the event where $\phi(\x)_u = \kappa, M_{uv}^i = 1$ and all other random variables (except for $\xi_{uv}$) take values such that, if $\xi_{uv} =+1$ then $\p_v \phi(\x)_u   \xi_{uv} M_{uv}^i  = h\round{\{\y_v\},  \kappa}$. On the other hand, if $\xi_{uv} =-1$ then $\p_v \phi(\x)_u   \xi_{uv} M_{uv}^i  = h\round{\{\y_v\},  -\kappa}$. Then, subtracting term 3 from term 1 (and factoring out the shared term) we find
\begin{align}
\begin{split}
    &\PR{\p_v \phi(\x)_u   \xi_{uv} M_{uv}^i  = h\round{\{\y_v\},  \kappa} | \phi(\x)_u = \kappa } \\ &- \PR{\p_v \phi(\x)_u   \xi_{uv} M_{uv}^i  = h\round{\{\y_v\},  -\kappa} | \phi(\x)_u = \kappa }
\end{split} \\
&=\PR{  \xi_{uv}  = +1 | H } - \PR{ \xi_{uv}  = -1 | H }  =0
\end{align}
since $\xi_{uv}$ is independent of $H$, and $\xi_{uv}= -1$ and $+1$ with equal probability.
\end{proof}
\end{lemma}

\end{document}